\documentclass[journal, twoside]{IEEEtran}

\pdfoutput=1

\usepackage{amssymb}
\usepackage{booktabs}     
\usepackage{multirow}
\usepackage{ulem}         
\usepackage{graphicx}
\usepackage{pifont}
\usepackage{array}
\usepackage{subfigure}    
\usepackage{hhline}
\usepackage{color}
\usepackage{epstopdf}
\usepackage{bbding}

\graphicspath{{figures/}}

\ifCLASSINFOpdf
\else
\fi
\usepackage{amsmath}
\usepackage{array}
\usepackage{url}

\newcommand{\PreserveBackslash}[1]{\let\temp=\\#1\let\\=\temp}
\newcolumntype{C}[1]{>{\PreserveBackslash\centering}p{#1}}
\newcolumntype{R}[1]{>{\PreserveBackslash\raggedleft}p{#1}}
\newcolumntype{L}[1]{>{\PreserveBackslash\raggedright}p{#1}}

\begin{document}
\title{Single MR Image Super-Resolution via Channel Splitting and Serial Fusion Network}
%

\author{Xiaole~Zhao,~Huali~Zhang,~Hangfei~Liu,~Yun~Qin,~Tao~Zhang,~Xueming~Zou
\thanks{This work has been submitted to the IEEE for possible publication. Copyright may be transferred without notice, after which this version may no longer be accessible.}
\thanks{X. Zhao is with the School of Life Science and Technology, University of Electronic Science and Technology of China (UESTC), Chengdu, Sichuan 611731, China (e-mail: zxlation@foxmail.com).}
\thanks{H. Zhang, H. Liu and Y. Qin are with the School of Life Science and Technology, University of Electronic Science and Technology of China (UESTC), Chengdu, Sichuan 611731, China.}
\thanks{T. Zhang is with the High Field Magnetic Resonance Brain Imaging Laboratory of Sichuan and Key Laboratory for Neuro Information of Ministry of Education, Chengdu, Sichuan 611731, China; He is also with the School of Life Science and Technology, University of Electronic Science and Technology of China (UESTC), Chengdu, Sichuan 611731, China (e-mail: taozhangjin@gmail.com).}
\thanks{X. Zou is with the School of Life Science and Technology, University of Electronic Science and Technology of China (UESTC), Chengdu, Sichuan 611731, China (mark.zou@alltechmed.com).}}

\markboth{}
{Zhao \MakeLowercase{\textit{et al.}}: Single MR Image Super-Resolution via Channel Splitting and Serial Fusion Network}
%



\maketitle

\begin{abstract}
Spatial resolution is a critical imaging parameter in magnetic resonance imaging (MRI). Acquiring high resolution MRI data usually takes long scanning time and would subject to motion artifacts due to hardware, physical, and physiological limitations. Single image super-resolution (SISR), especially that based on deep learning techniques, is an effective and promising alternative technique to improve the current spatial resolution of magnetic resonance (MR) images. However, the deeper network is more difficult to be effectively trained because the information is gradually weakened as the network deepens. This problem becomes more serious for medical images due to the degradation of training examples. In this paper, we present a novel channel splitting and serial fusion network (CSSFN) for single MR image super-resolution. Specifically, the proposed CSSFN network splits the hierarchical features into a series of subfeatures, which are then integrated together in a serial manner. Thus, the network becomes deeper and can deal with the subfeatures on different channels discriminatively. Besides, a dense global feature fusion (DGFF) is adopted to integrate the intermediate features, which further promotes the information flow in the network. Extensive experiments on several typical MR images show the superiority of our CSSFN model over other advanced SISR methods.
\end{abstract}


%
\IEEEpeerreviewmaketitle

\section{Introduction}
%
%
%
%
\IEEEPARstart{M}{agnetic} resonance imaging (MRI) is an important and widely used tool for diagnosis and image-guided therapeutics. High resolution (HR) magnetic resonance (MR) images are usually preferred in clinical practice due to more clear image structure and texture details, as well as the benefits to subsequent analysis and processing \cite{Carmia2006Resolution,Oktay2016Multi}. However, the acquisition of HR images is constrained by hardware, physical and physiological factors, and increasing the spatial resolution of MR images typically reduces the signal noise ratio (SNR) and/or increases imaging time \cite{Plenge2012Super}, which further increases the risk of MR images affected by motion artifacts.

\begin{figure}[t]
  \centering
  \subfigure[Validation curves]{\label{fig1a}
  \begin{minipage}[t]{0.23\textwidth}
    \centering
    \includegraphics[scale = 0.28]{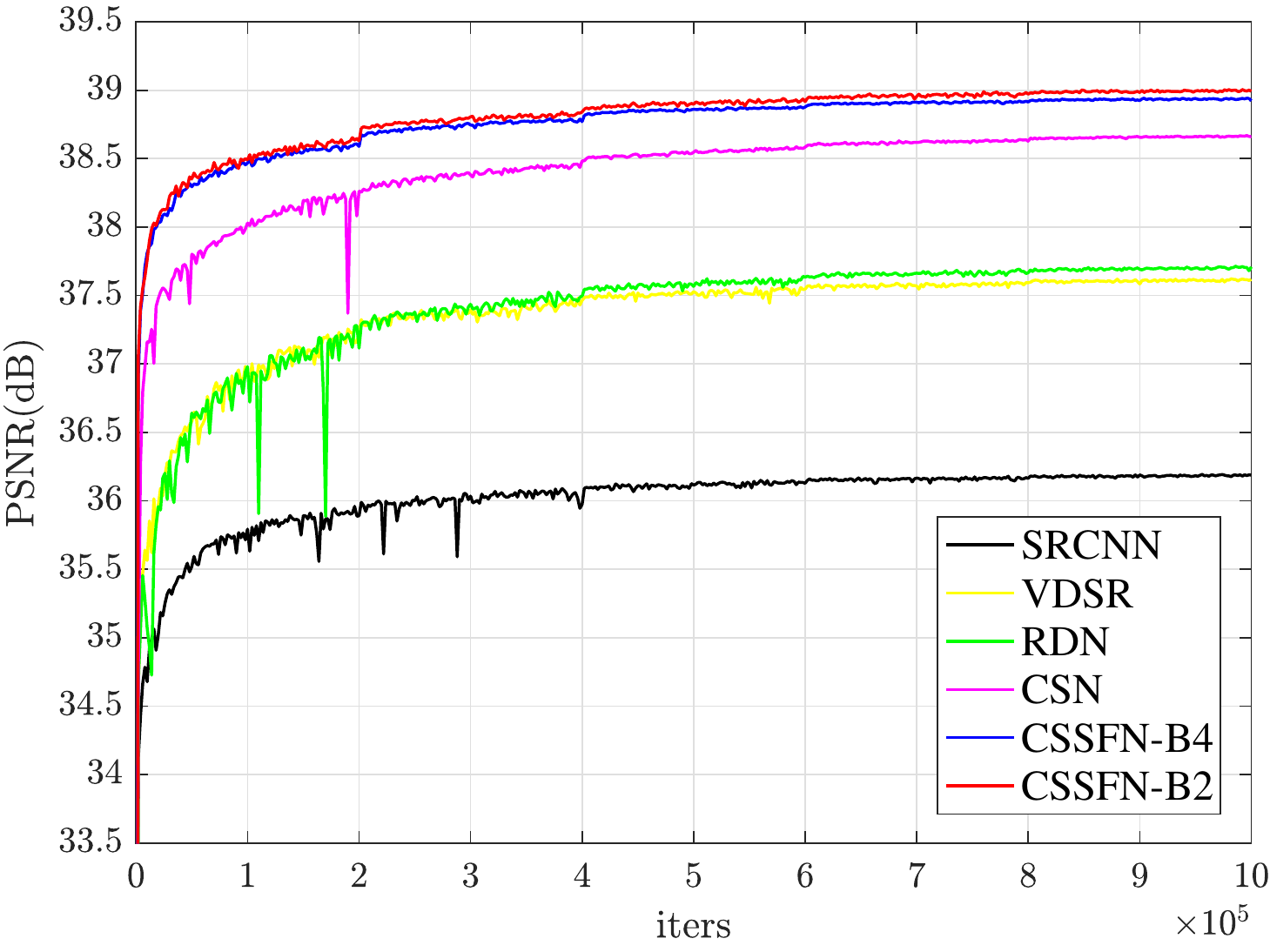}
  \end{minipage}}
  \subfigure[Testing results]{\label{fig1b}
  \begin{minipage}[t]{0.23\textwidth}
    \centering
    \includegraphics[scale = 0.28]{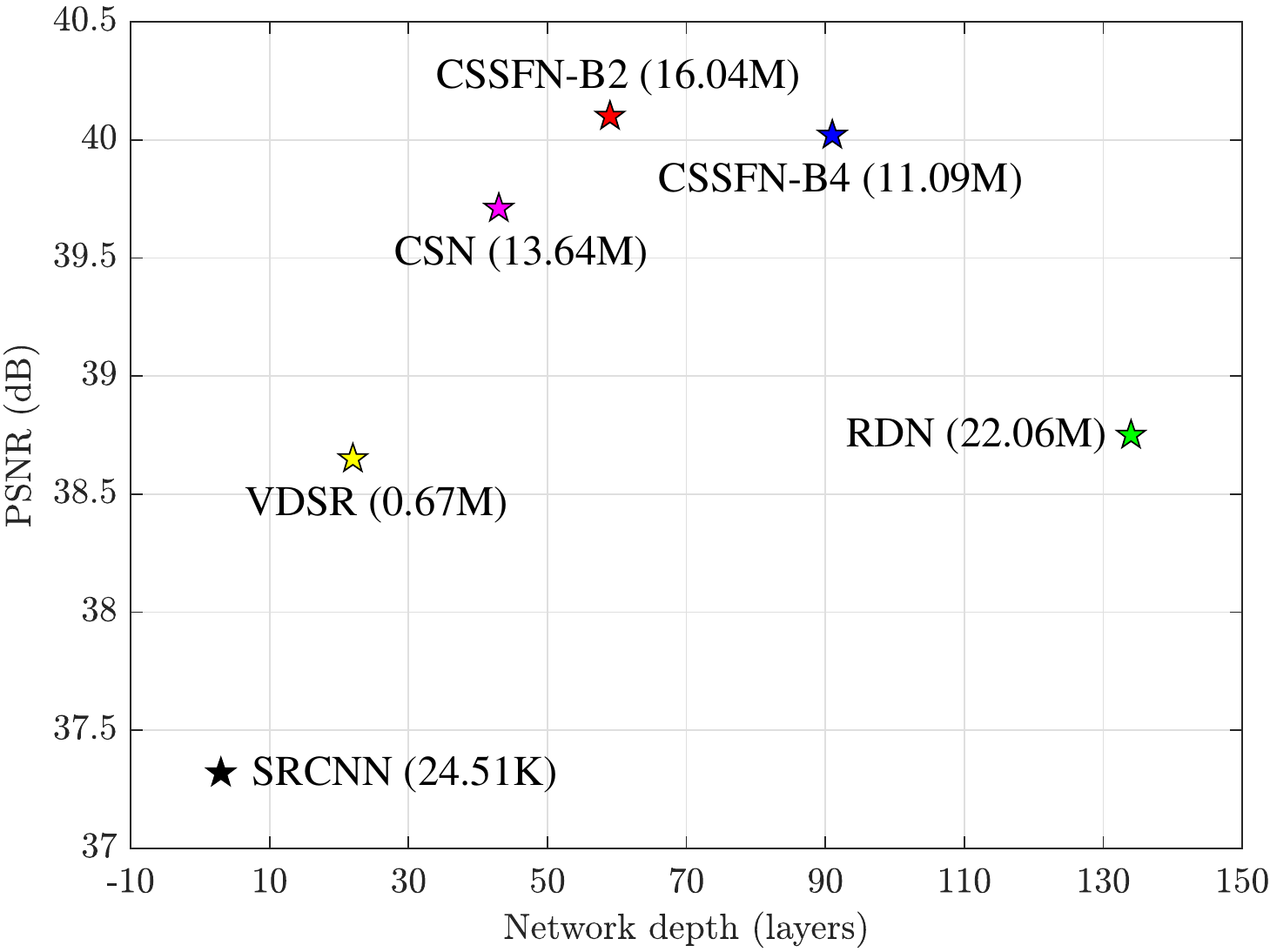}
  \end{minipage}}
  \vspace{-2mm}
  \caption{The PSNR performance versus  model scale (model parameters and network depth). The results are evaluated on T2 MR images of IXI dataset for SR$\times$2, and ``B2'' and ``B4'' imply the number of subfeatures when performing channel splitting. It can be seen that our CSSFN models have a better tradeoff between SR performance and model scale.}\label{fig1}
\end{figure}

Image super-resolution (SR) is an effective and cost efficient alternative technique to increase the spatial resolution of MR images, which aims at inferring a HR image from one or more low resolution (LR) images. Up to now, many SR algorithms have been investigated and proposed for both natural images and medical images, e.g., interpolation-based and edge-guided methods \cite{Keys1981Cubic,Gottlieb1997On,Li2001New,Sun2008Image}, modeling and reconstruction based methods \cite{Irani1991Improving,Stark1989High,Peled2001Super,Greenspan2002MRI}, example learning based methods \cite{Freeman2002Example,Kim2013Example}, and dictionary learning and sparse representation methods \cite{Yang2008Image,Yang2012Coupled,Rueda2013Single,Wang2014Sparse} etc. However, the performance of these conventional methods is essentially limited because they apply inadequate additional information and models with limited representational capacity to solve the notoriously challenging ill-posed inverse problem of image SR tasks \cite{Zhao2018Channel,Yang2018Deep}.

\begin{figure*}[t]
  \centering
  \includegraphics[width = \textwidth]{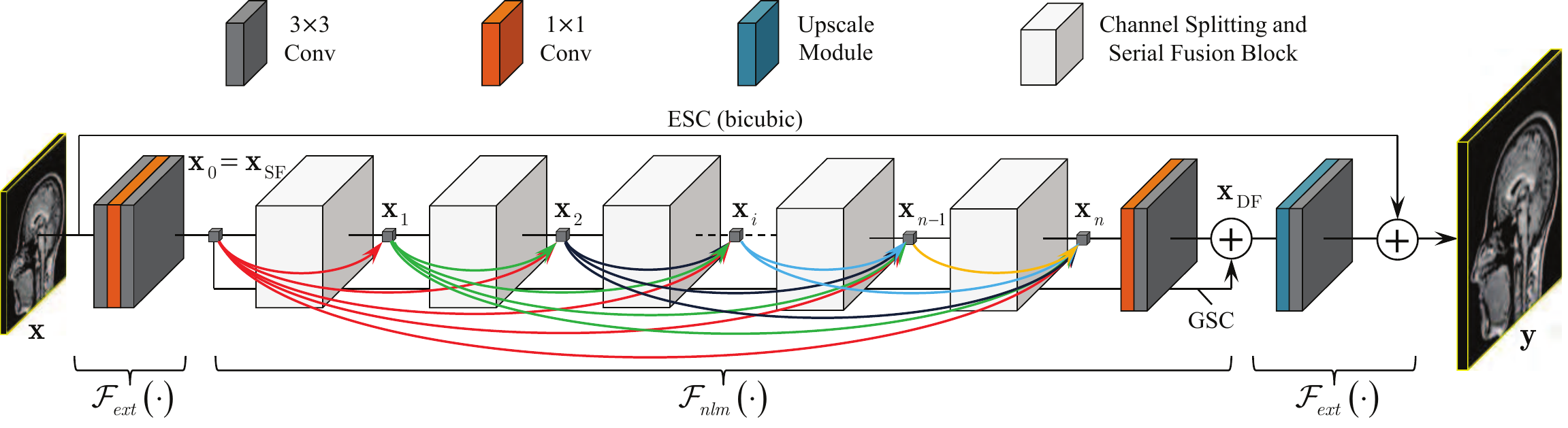}   \\
  \vspace{-2mm}
  \caption{The overall architecture of the proposed channel splitting and serial fusion network (CSSFN). The symbol ``$\boldsymbol{+}$'' indicates element-wise summation between two tensors with the same shape. GSC and ESC denote global skip connection and external skip connection, respectively. The hierarchical features are integrated together in a dense learning manner \cite{Huang2016Densely,Tong2017Image}. We term this as dense global feature fusion (DGFF).}\label{fig2}
\end{figure*}

In recent years, deep learning \cite{Lecun2015Deep} based single image super resolution (SISR) methods have demonstrated great superiority over conventional SR methods. A pioneering work that uses convolutional neural networks (CNNs) \cite{Lecun1998Gradient} to deal with SISR is the super-resolution convolutional neural network (SRCNN) \cite{Dong2016Image,Dong2016Accelerating}. It implicitly learns an end-to-end mapping function between LR and HR images by utilizing a fully-convolutional network. Subsequently, many more advanced SISR techniques based on deep CNNs were proposed. Some typical examples are DRCN \cite{Kim2016Deeply}, DRRN \cite{Tai2017Image}, VDSR \cite{Kim2016Accurate}, MemNet \cite{Tai2017Memnet}, ESPCNN \cite{Shi2016Real}, SRResNet \cite{Ledig2016Photo}, EDSR/MDSR \cite{Lim2017Enhanced}, RDN \cite{Zhang2018Residual}, CMSCN \cite{Hu2018Single} and RCAN \cite{Zhang2018Image} etc. These methods have overwhelming advantages over traditional methods and greatly promote the best state of SISR performance. However, they are mainly aimed at the SISR task of natural images, instead of medical images (or more specifically, MR images). Thus, they may be unsuitable for solving medical image SR tasks due to the degradation of training examples \cite{Zhao2018Channel}, although they have excellent performance on natural images.

Some deep learning based methods specializing in the SISR tasks of medical images have also emerged due to the tremendous success of deep learning techniques in computer vision and pattern recognition \cite{Pham2017Brain,Chen2018Brain,Chen2018Efficient,You2018CT}. These methods utilize relatively shallow network structures to process medical images, e.g., Pham \textit{et al}. \cite{Pham2017Brain} presented an algorithm for brain MR images SR according to SRCNN \cite{Dong2016Image}. Despite they extend it to 3D cases (named SRCNN3D), the entire network is very shallow and the representational ability of the model is relatively limited, resulting in unsatisfactory SR performance.

\begin{figure*}[t]
  \centering
  \includegraphics[width = \textwidth]{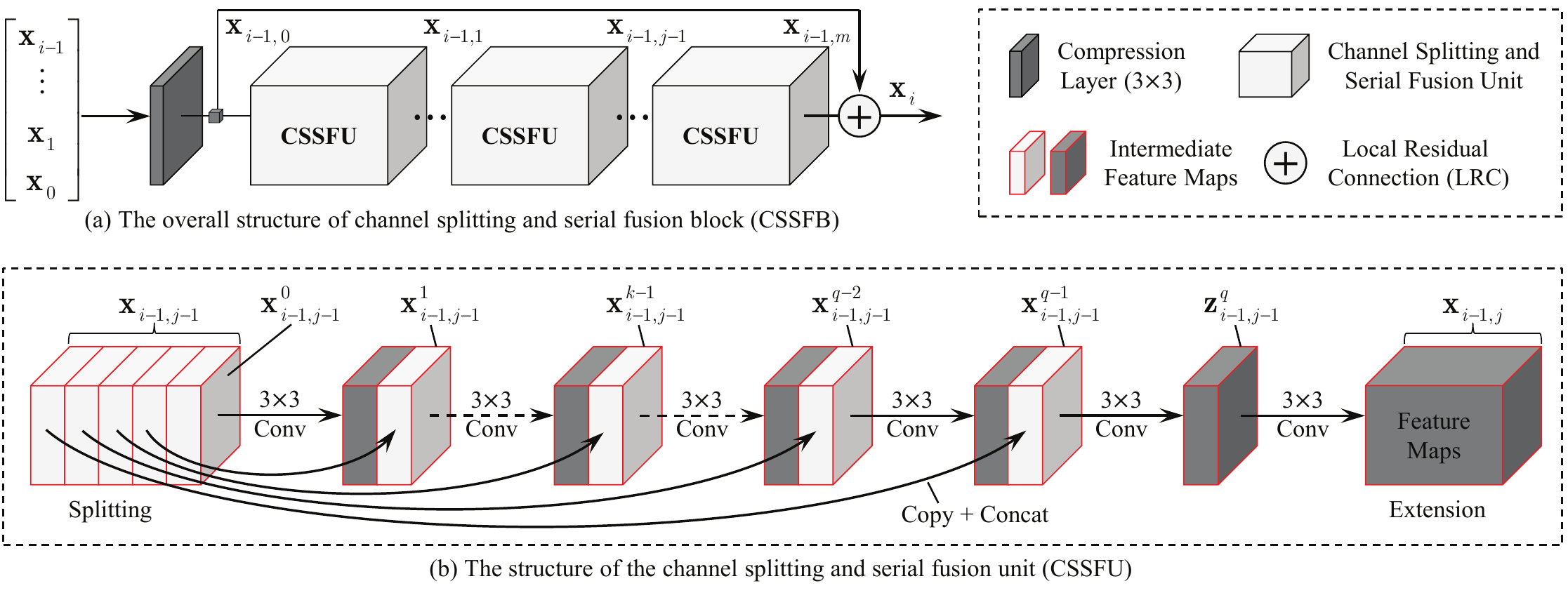}   \\
  \vspace{-2mm}
  \caption{The overall structure of the basic building block, which consists of $m$ stacked channel splitting and serial fusion units (CSSFU). (a) Each CSSFB also has a short skip connection (SSC) to form local residual learning. (b) The input feature of each CSSFU is split into $q$ subfeatures. Cuboids in light gray imply subfeatures from channel splitting of the input feature, and those in dark gray denote the subfeatures produced by the 3$\times$3 conv layer.}\label{fig3}
\end{figure*}

The depth of deep CNNs is of crucial importance for the task of image SR \cite{Zhang2018Image}, and usually defined as the longest path from the input to the output \cite{Zhao2018Channel,Tai2017Image,Hu2018Single}. However, the deeper networks are more difficult to be effectively and fully trained, especially with medical images due to the degeneracy of training samples \cite{Zhao2018Channel}. Actually, it is verified that the original EDSR model \cite{Lim2017Enhanced} with about 43M  model parameters and 70 layers of depth is difficult to be well-trained with 2D proton density (PD) images \cite{Zhao2018Channel}. Although Zhao \textit{et al}. \cite{Zhao2018Self} trained the EDSR model \cite{Lim2017Enhanced} using T1-weighted magnetization-prepared rapid gradient echo (MP-RAGE) images, the reported results are not satisfactory. This is probably because that the EDSR model, which has enormous parameters and very deep network structure, is not adequately well-trained with ``good'' training samples. In this regard, whether deeper networks are capable of further contributing to improve the performance of medical image SR and how to construct trainable networks with much deeper structure for medical images remain to be explored.

A recent work \cite{Zhao2018Channel} has alleviated the dilemma between the trainability and the performance of deep CNN models for MR image SR to some extent. It presented an effective manner to deepen the network but without significant increase in the number of model parameters, i.e., channel splitting. The model proposed by \cite{Zhao2018Channel}, however, is a kind of multistream structure and the multiple information branches are formed by channel splitting instead of the reuse of preceding features \cite{Hu2018Single,Zhao2017Deep}. This multistream structure implies that the information flow in the network is \textit{locally parallel}. In this work, we present a serial information fusion mechanism for channel splitting. The proposed model, which we term as channel splitting and serial fusion network (CSSFN), first splits the hierarchical features into a series of subfeatures and then integrates them together in a serial manner. Different from the CSN \cite{Zhao2018Channel}, the proposed CSSFN network is a single-branch structure. Thus, it can reach a deeper structure (up to 90 layers) than the CSN model with fewer model parameters (Fig.\ref{fig1}). On the other hand, channel splitting also allows the proposed CSSFN model to deal with features on different channels discriminatorily. But unlike the CSN model, each subfeature in our framework has different depth in the nonlinear mapping process.

To alleviate the instability of model training caused by the single-branch architecture and the increase in network depth, we employ a dense global feature fusion (DGFF) strategy to enhance the information flow in the network. The proposed CSSFN model consists of a series of building blocks, each of which contains a battery of channel splitting and serial fusion units (CSSFU). Each building block has one or more inter-block connections to all subsequent building blocks, thus propagating its own local features to all successors. Although the DGFF increases the model parameters to some extent, our CSSFN model still have moderate model parameters compared with EDSR \cite{Lim2017Enhanced}, RDN \cite {Zhang2018Residual} and even CSN \cite{Zhao2018Channel}.

The remainder of this work is organized as follows. We first present some previous work related the proposed method in section \ref{sec:relatedwork}. Then, the proposed method is illustrated in detail in section \ref{sec:proposedmethod} and the experimental results are presented in section \ref{sec:experiments}. Finally, we conclude the paper in section \ref{sec:conclusion}.

\begin{table*}[t]
  \centering
  \caption{The statistics of network depth and model parameters when $c = 256$ and $n = m = 4$. Here $ic$ denotes the number of input channels.}
  \vspace{-2mm}
  \begin{tabular}{C{1.4cm}|C{0.9cm}|C{0.9cm}|C{0.9cm}|C{0.9cm}|C{0.9cm}|C{0.9cm}||C{0.9cm}|C{0.9cm}|C{0.9cm}|C{0.9cm}|C{0.9cm}|C{0.9cm}}
  \toprule
    Execution  & \multicolumn{6}{c||}{Pure 2D ($ic = 1$)} & \multicolumn{6}{c}{Pseudo 3D ($ic = 96$)} \\
    \hhline{-------||------}
    Subfeature   & \multicolumn{3}{c|}{$q = 2$}  & \multicolumn{3}{c||}{$q = 4$} & \multicolumn{3}{c|}{$q = 2$} & \multicolumn{3}{c}{$q = 4$} \\
    \hhline{-------||------}
    $r$      &$\times$2&$\times$3&$\times$4&$\times$2&$\times$3&$\times$4&$\times$2&$\times$3&$\times$4&$\times$2&$\times$3&$\times$4 \\
    \hhline{-------||------}
    $D$      &   59    &   59    &   60    &   91    &    91   &     92  &   59    &   59    &    60   &     91  &    91   &    92    \\
    \hhline{-------||------}
    $P$      & 16.40M  & 19.35M  & 18.76M  & 11.09M  &  14.04M &  13.45M & 16.84M  & 19.79M  & 19.20M  &  11.53M &  14.48M &  13.89M  \\
    \bottomrule
  \end{tabular}
  \label{tab:1}
\end{table*}

\section{Related Work}
\label{sec:relatedwork}
\subsection{MR Image Super-Resolution}
The purpose of MR image SR methods is to overcome the hardware limitations and meet the clinical needs of imaging procedures by reconstructing HR images from LR acquisitions using post-processing methods. These SR methods could have strong impacts on structural MRI when focusing on cortical surface or fine-scale structure analysis \cite{Pham2017Brain}. The application of SR methods to MR images initially focuses on multiple image super resolution (MISR), e.g., \cite{Peled2001Super,Greenspan2002MRI,Shilling2009A}. However, MISR methods usually need calibration and fusion between multiple LR images, which is a challenging problem in itself \cite{Zhao2018Channel}.

SISR can avoid the difficulty of calibration and fusion faced by MISR, where only one LR image is required to predicate its HR counterpart, e.g., \cite{Rueda2013Single,Wang2014Sparse,Manj車n2010MRI,Rousseau2008Brain}. A major problem with SISR methods is that there is limited extra information available for HR image reconstruction. Subsequently, some SR methods based on traditional machine learning, e.g., sparse representation \cite{Rueda2013Single,Wang2014Sparse}, example learning \cite{Trinh2014Novel,Hu2016Single}, as well as compressive sensing \cite{Roohi2012Super} etc., have emerged. However, the limited representational capability of these SR methods makes them unable to accurately reflect the highly nonlinear mapping between LR and HR images. Recently, more advanced SISR methods based on deep learning \cite{Lecun2015Deep} have also been applied to MR image SR tasks \cite{Oktay2016Multi,Zhao2018Channel,Pham2017Brain,Chen2018Brain,Chen2018Efficient,Zhao2018Self}, which have greatly promoted the performance of SR technologies for medical images or, more specifically, MR images.

\subsection{Channel Discrimination}
The feature mappings on different channels of deep CNNs have different types of information and different impacts on the performance of deep models \cite{Zhang2018Image,Hu2018Channel}, and it is reasonable to deal with the feature mappings discriminatorily. One typical way of channel discrimination is attention mechanism, which is broadly viewed as a tool to bias the allocation of available processing resources towards the most informative components of the input signal \cite{Hu2017Squeeze}. In recent years, it has been introduced to deep neural networks (DNNs) to boost the performance of deep models, such as image generation \cite{Mansimov2016Generating}, image captioning \cite{Xu2015Show,Chen2017SCACNN}, image classification \cite{Hu2017Squeeze,Wang2017Residual} and image restoration \cite{Zhang2018Image,Wang2018Recovering,Hu2018Channel,Cheng2018SESR}. These methods have further improved the best state of related fields. For instance, the residual channel attention network (RCAN) \cite{Zhang2018Image} pushed the state-of-the-art SR performance forward on natural images, with an extremely deep network structure (over 400 layers).

However, few works have been conducted to investigate the effect of channel discrimination for low-level computer vision tasks in medical image processing community (e.g., MR image SR). In this respect, a representative work for single MR image SR is the CSN network \cite{Zhao2018Channel}, where channel discrimination is achieved by channel splitting and the merge-and-run mapping between different branches \cite{Zhao2017Deep}. This model adopted a parallel two-way channel splitting strategy to handle the hierarchical features on different channels, which limited the network depth to some extent. Inspired by channel discrimination mechanism and increasing the network depth, we integrate the subfeatures into a single branch in a serial manner (Fig.\ref{fig3}(b)). Thus, the network becomes deeper and thinner (if we constrain the width of subfeatures), which is analogous to stretching a rubber band. Despite the single branch, the serial fusion retains the channel discrimination ability of the network, because the subfeatures have different depths in the processing.

\subsection{Hierarchical Feature Fusion}
The notorious problems of gradient vanishing and weakened information flow becomes more obvious as the network depth increases \cite{Zhang2018Image,Hu2018Channel}, which hinders the training of deep models seriously. Unfortunately, the degradation of training samples will further aggravate the difficulty of training deep models in the context of medical images \cite{Zhao2018Channel}. In order to promote the information flow in the network and improve the trainability of the model, many recent works have been devoted to resolving these problems. A popular  method is to fuse the hierarchical features through skip connections, e.g., DenseNet \cite{Huang2016Densely} helps to explore new feature maps, and ResNet \cite{He2016Deep,He2015Identity} contributes to the reuse of the preceding features. The basic idea of fusing hierarchical features by residual learning and dense learning is also widely applied to many CNN-based methods, e.g., \cite{Kim2016Accurate,Ledig2016Photo,Lim2017Enhanced,Zhang2018Residual, Zhang2018Image,Hu2018Channel,Tong2017Image,Wang2018Deep}, to build very wide and deep networks for performance improvement.

Since most recent CNN-based SR models are modular, the hierarchical feature fusion can be divided into local feature fusion (LFF) and global feature fusion (GFF), which integrate intra-block and inter-block features, respectively. The LFF is conducive to learning more effective hierarchical features and stabilizing model training \cite{Zhang2018Residual}, while the GFF enables short paths to be built from high-level features to low-level features directly and further alleviate the problem of gradient vanishing for training very deep networks \cite{Tong2017Image}. In the proposed CSSFN model, the local features are fused together simply by a local residual connection (LRC) \cite{Zhang2018Residual}, which is also known as short skip connection (SSC) \cite{Zhang2018Image} or shortcut connection \cite{He2016Deep,He2015Identity} (Fig.\ref{fig3}(a)). However, the serial fusion of subfeatures in the CSSFU can be viewed as a manner of partially dense learning where the subfeatures are ``densely'' connected to subsequent layers (Fig.\ref{fig3}(b)). Besides, we also use a dense global feature fusion (DGFF) for effective feature exploitation and important information preservation (Fig.\ref{fig2}). More importantly, it helps to relieve the instability of model training caused by the increase in network depth and the decrease in network width.

\begin{figure}[t]
  \centering
  \includegraphics[width = 0.48\textwidth]{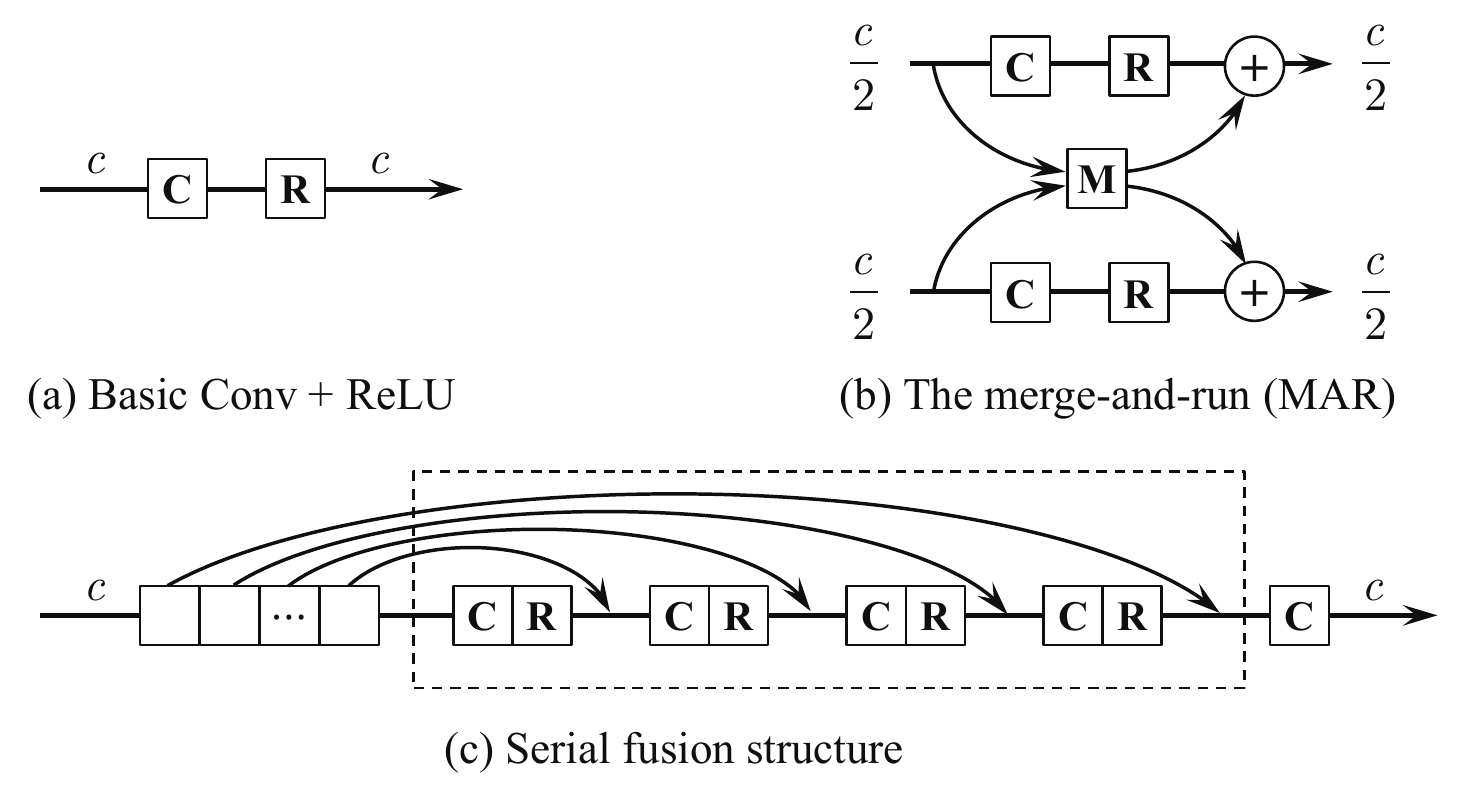}   \\
  \vspace{-4mm}
  \caption{Three structures of stage mapping for comparing branch information fusion (BIF). ``C'', ``R'' and ``$\boldsymbol{+}$'' represent Conv, ReLU, and skip connection respectively. (a) The basic Conv + ReLU structure. (b) The merge-and-run structure \cite{Zhao2017Deep} with channel splitting \cite{Zhao2018Channel}. The number of branches is set to 2 for display purposes. (c) The proposed serial fusion structure. The number of channels in the dashed box can be adjusted accordingly.}
  \label{fig4}
\end{figure}

\begin{figure}[t]
  \centering
  \subfigure[Pure 2D execution ($ic = 1$)]{\label{PC-C1}
  \begin{minipage}[t]{0.23\textwidth}
    \centering
    \includegraphics[scale = 0.28]{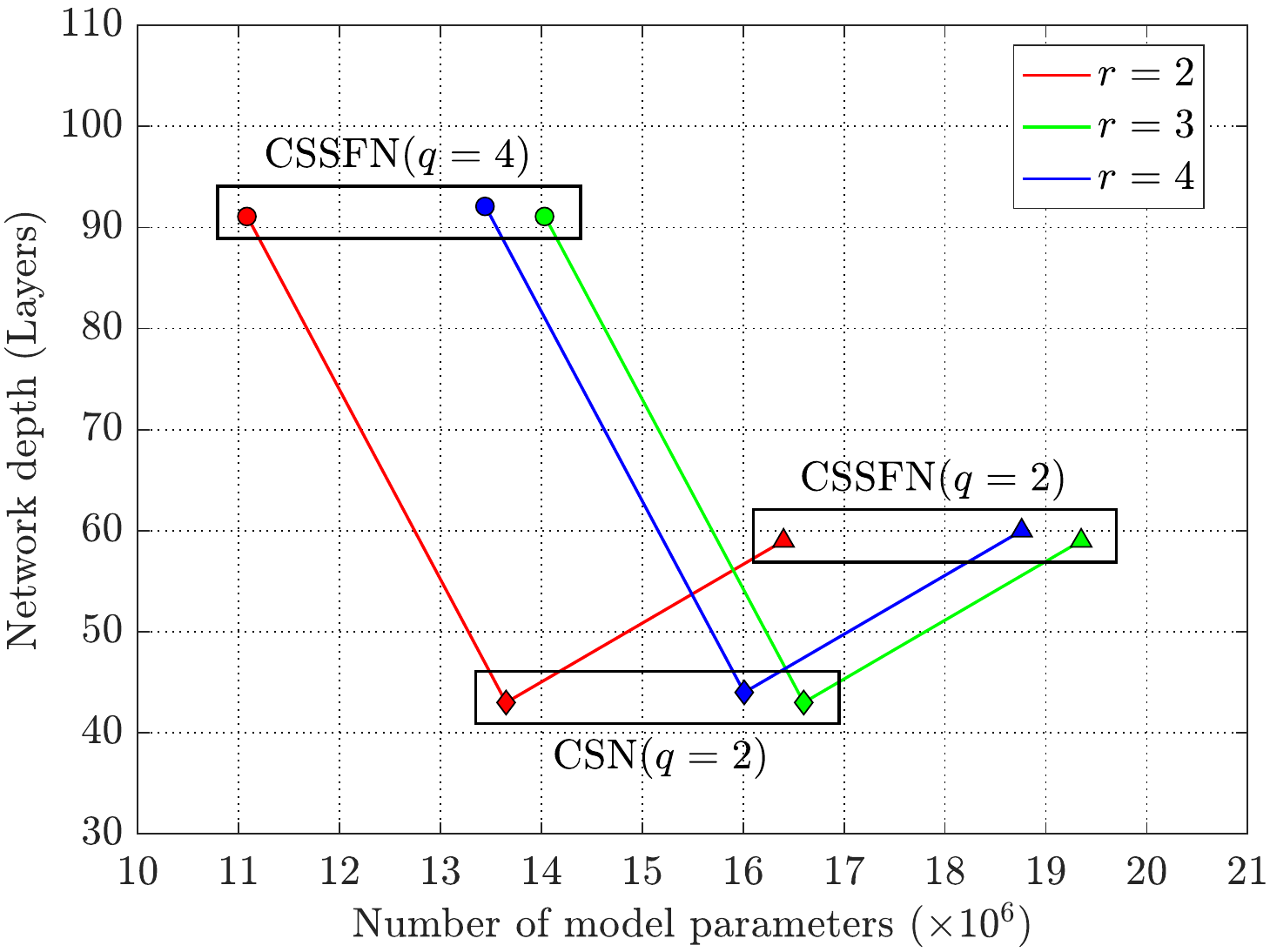}
  \end{minipage}}
  \subfigure[Pseudo 3D execution ($ic = 96$)]{\label{PC-C96}
  \begin{minipage}[t]{0.23\textwidth}
    \centering
    \includegraphics[scale = 0.28]{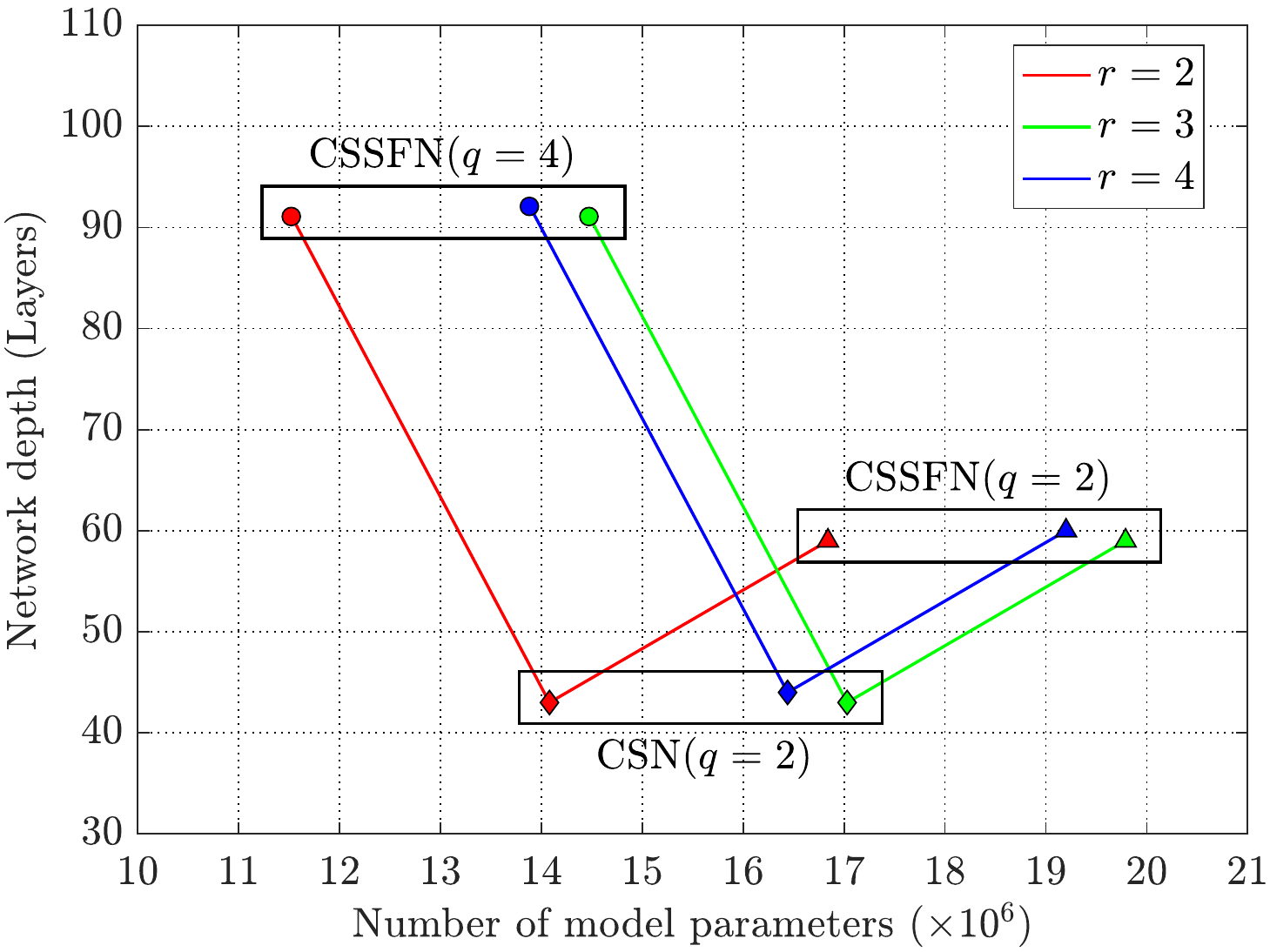}
  \end{minipage}}
  \vspace{-2mm}
  \caption{Comparison of network depth and model parameters between the CSN \cite{Zhao2018Channel} and the proposed CSSFN. For all compared models, we set $m = n = 4$ and $c = 256$. The symbols $\vartriangle$ and $\circ$ represent CSSFN with $q = 2$ and $q = 4$ respectively, and $\diamond$ denotes the CSN model with 2 branches.}
  \label{fig:param-compare}
\end{figure}

\begin{table*}[t]
  \centering
  \caption{Ablation investigation of different global feature fusion (GFF) and branch information fusion (BIF) methods. All the models are trained on $\mathcal{D}$(T2, BD) for one million iterations and tested on $\mathcal{T}$(T2, BD). SF denotes serial fusion ($c = 256, m = n = q = 4$).}
  \vspace{-2mm}
  \begin{tabular}{C{1.1cm}|C{1.4cm}|C{1.2cm}|C{1.2cm}|C{1.2cm}|C{1.2cm}|C{1.2cm}|C{1.2cm}|C{1.2cm}|C{1.2cm}|C{1.2cm}}
    \toprule
    \multirow{2}{*}{GFF} & CGFF & {\tiny \XSolid} & $\surd$         & {\tiny \XSolid} & {\tiny \XSolid} & {\tiny \XSolid} & $\surd$         & $\surd$ & {\tiny \XSolid} & {\tiny \XSolid} \\
    \cmidrule{2-11}
                         & DGFF & {\tiny \XSolid} & {\tiny \XSolid} & $\surd$         & {\tiny \XSolid} & {\tiny \XSolid} & {\tiny \XSolid} & {\tiny \XSolid} & $\surd$ & $\surd$ \\
    \midrule
    \multirow{2}{*}{BIF} & MAR  & {\tiny \XSolid} &  {\tiny \XSolid}& {\tiny \XSolid} & $\surd$         & {\tiny \XSolid} & $\surd$         & {\tiny \XSolid} & $\surd$ & {\tiny \XSolid} \\
    \cmidrule{2-11}
                         & SF   & {\tiny \XSolid} & {\tiny \XSolid} & {\tiny \XSolid} & {\tiny \XSolid} & $\surd$         & {\tiny \XSolid} & $\surd$  & {\tiny \XSolid} & $\surd$ \\
    \midrule
    \multirow{2}{*}{$r=2$} & PSNR (dB) & 39.90  & 39.95  & 39.88  & 39.66  & 40.01  & 39.66  & 40.03  & 39.68  & 40.05  \\
              & SSIM\ \ \ \ \ \ \ \ \  & 0.9867 & 0.9868 & 0.9866 & 0.9862 & 0.9869 & 0.9861 & 0.9869 & 0.9863 & 0.9870 \\
    \bottomrule
  \end{tabular}
  \label{tab:2}
\end{table*}

\section{Proposed Method}
\label{sec:proposedmethod}
\subsection{Network Architecture}
In this paper, we focus on the task of single 2D MR image super-resolution. Given a LR image $\mathbf{x} \in \mathbb{R}^{h \times w}$, the target here is to recover a HR image $\mathbf{y} \in \mathbb{R}^{(r \cdot h) \times (r \cdot w)}$ that corresponds to the LR input $\mathbf{x}$, where $r$ is the scaling factor. The overall architecture of the proposed CSSFN model is outlined in Fig.\ref{fig2} and Fig.\ref{fig3}, which consists of three typical parts, i.e., shallow feature extraction, nonlinear mapping from shallow features to deep features and HR image recovery. As investigated in \cite{Zhao2018Channel}, we extract the shallow features by two 3$\times$3 conv layers with a 1$\times$1 conv layer in the middle. Denote $\mathcal{F}_{ext}(\cdot)$ as the corresponding mapping function of the entire shallow feature extraction stage, then the extracted shallowed features $\mathbf{x}_{\text{SF}}$ can be represented as:
\begin{equation}
\label{SFE}
  \mathbf{x}_{\text{SF}} = \mathcal{F}_{ext}(\mathbf{x}),
\end{equation}
where $\mathbf{x}$ denotes the original LR input. Next, $\mathbf{x}_{\text{SF}}$ is fed into the nonlinear mapping, which contains a series of stacked building blocks. The entire nonlinear mapping process can be expressed as follows:
\begin{equation}
\label{NLM}
  \mathbf{x}_{\text{DF}} = \mathcal{F}_{nlm}(\mathbf{x}_{0}),
\end{equation}
where $\mathbf{x}_{0} = \mathbf{x}_{\text{SF}}$ is the extracted shallow features and $\mathcal{F}_{nlm}(\cdot)$ is the function corresponding to the entire nonlinear mapping process. To make more full use of the hierarchical features and further stabilize model training, we also utilize the global feature fusion (GFF) \cite{Zhao2018Channel,Zhang2018Residual} to integrate these intermediate features. Unlike \cite{Zhao2018Channel} and \cite{Zhang2018Residual}, we fuse the hierarchical features in a dense learning manner \cite{Tong2017Image}, instead of concatenating all the inter-block features together and then fuse them through a 1$\times$1 conv layer. Thus, the input to the $i$-th building block is the concatenation of the output feature maps of all preceding blocks, i.e., $[\mathbf{x}_{i-1}, \ldots, \mathbf{x}_{1}, \mathbf{x}_{0}]$, where $[\ldots]$ denotes the concat operation along channel direction. Assuming that the mapping function of the $i$-th building block is $\mathcal{F}_{b}^{i}(\cdot)$, then we have:
\begin{equation}\label{CSSFB}
  \mathbf{x}_{i} = \mathcal{F}_{b}^{i}([\mathbf{x}_{i-1}, \ldots, \mathbf{x}_{1}, \mathbf{x}_{0}]),\ \ i = 1,2,\ldots,n,
\end{equation}
where $n$ is the number of building blocks in the network. Each block is connected to all preceding and subsequent blocks, and therefore facilitates the information propagation of the entire network. Iteratively, we can obtain the final output of all these stacked blocks:
\begin{equation}
\label{NLM-CSSFB}
  \mathbf{x}_{n} = \mathcal{F}_{b}^{n}([\mathbf{x}_{n-1}, \ldots, \mathbf{x}_{1}, \mathbf{x}_{0}]).
\end{equation}
Then, $[\mathbf{x}_{n},\ldots, \mathbf{x}_{0}]$ is further integrated as the deep feature $\mathbf{x}_{\text{DF}}$ through a 1$\times$1 conv layer and a 3$\times$3 conv layer, followed by a global skip connection (GSC) \cite{Zhao2018Channel,Kim2016Accurate,Lim2017Enhanced,Zhang2018Residual}:
\begin{equation}\label{NLM-LAST}
  \mathbf{x}_{\text{DF}} = \mathbf{x}_{0} + \mathcal{F}_{c}([\mathbf{x}_{n},\ldots,\mathbf{x}_{0}]),
\end{equation}
where $\mathcal{F}_{c}(\cdot)$ corresponds to the mapping function of the two conv layers, as shown in Fig.\ref{fig2}. Subsequently, the deep feature $\mathbf{x}_{\text{DF}}$ is used to recover the HR image $\mathbf{y}$ by the reconstruction sub network:
\begin{equation}
\label{REC}
  \mathbf{y} = \mathcal{F}_{rec}(\mathbf{x}_{\text{DF}}) = \mathcal{F}_{up}(\mathbf{x}_{\text{DF}}) + \hat{\mathbf{x}},
\end{equation}
where $\mathcal{F}_{up}(\cdot)$ represents the mapping function of the upscale module followed by a 3$\times$3 conv layer, and $\hat{\mathbf{x}}$ is the (bicubic) interpolated version of $\mathbf{x}$. This is termed as an external skip connection (ESC) in \cite{Zhao2018Channel}, which approximates the residual between the original input and the final output of the network by interpolation \cite{Kim2016Accurate}, and further contributes to stabilizing the training process.

We adopt $L_1$ loss as the training objective. Given a training dataset $\mathcal{D}=\{\mathbf{x}^{(i)}, \mathbf{y}^{(i)}\}_{i=1}^{|\mathcal{D}|}$, where $|\mathcal{D}|$ denotes the number of training examples in $\mathcal{D}$. The loss function is expressed as:
\begin{equation}
\label{OBJ}
  L(\boldsymbol{\theta}) = \frac{1}{|\mathcal{D}|}\sum_{i = 1}^{|\mathcal{D}|}||\mathbf{y}^{(i)} - \mathcal{F}_{net}(\mathbf{x}^{(i)}; \boldsymbol{\theta})||_{1}.
\end{equation}
Here, $\mathcal{F}_{net}(\cdot)$ denotes the function corresponding to the entire CSSFN network, and $\boldsymbol{\theta}$ is the set of model parameters.

\subsection{Channel Splitting and Serial Fusion Block (CSSFB)}
\label{subsec:cssfb}
The structure of the proposed channel splitting and serial fusion block (CSSFB) is outlined in Fig.\ref{fig3}(a). At the beginning of each CSSFB, there is a 3$\times$3 channel compression layer, which is used to reduce the feature dimension of the input tensor to a predefined value $c$. According to (\ref{CSSFB}), we have:
\begin{equation}
\label{COMPRESSION}
  \mathbf{x}_{i - 1, 0} = \mathcal{H}_{1}^{c}([\mathbf{x}_{i - 1}, \ldots, \mathbf{x}_{1}, \mathbf{x}_{0}]),\ \ \ \ i = 1,2, \ldots, n,
\end{equation}
where $ \mathcal{H}_{1}^{c}(\cdot)$ represents a convolution layer with 1$\times$1 kernel size and $c$ output channels. This implies that the feature map $\mathbf{x}_{i - 1, 0}$ has $c$ channels.

Subsequently, a series of stacked channel splitting and serial fusion units (CSSFUs) form the main part of the CSSFB, as shown in Fig.\ref{fig3}(a). Let's denote the function of the $j$-th CSSFU as $\mathcal{F}_{u}^{j}(\cdot)$, which we will describe in \ref{subsec:dcsu} in detail. Then we have the following equation for this CSSFU:
\begin{equation}
\label{DCSU}
  \mathbf{x}_{i-1,j} = \mathcal{F}_{u}^{j}(\mathbf{x}_{i-1,j-1}),\ \ \ \ j = 1,2,\ldots,m,
\end{equation}
where $m$ is the number of CSSFUs in each CSSFB. We can also get the output of the last CSSFU $\mathbf{x}_{i-1,m}$ iteratively:
\begin{equation}
\label{DCSU-LAST}
\begin{aligned}
  \mathbf{x}_{i-1,m} &= \mathcal{F}_{u}^{m}(\mathbf{x}_{i-1,m-1})\\
                     &= \mathcal{F}_{u}^{m}(\mathcal{F}_{u}^{m-1}(\cdots \mathcal{F}_{u}^{1}(\mathbf{x}_{i-1,0}) \cdots)).
\end{aligned}
\end{equation}
Local residual learning (LRL) \cite{Zhao2018Channel,Lim2017Enhanced,Zhang2018Residual,Zhang2018Image,Hu2018Channel} is another manner to stabilize model training. We also introduce LRL into the proposed CSSFB modules, so the final output of the $i$-th CSSFB can be expressed as:
\begin{equation}
\label{LRL}
  \mathbf{x}_{i} = \mathbf{x}_{i-1,0} + \mathbf{x}_{i-1,m}.
\end{equation}

\begin{figure}[t]
  \centering
  \includegraphics[width = 0.48\textwidth]{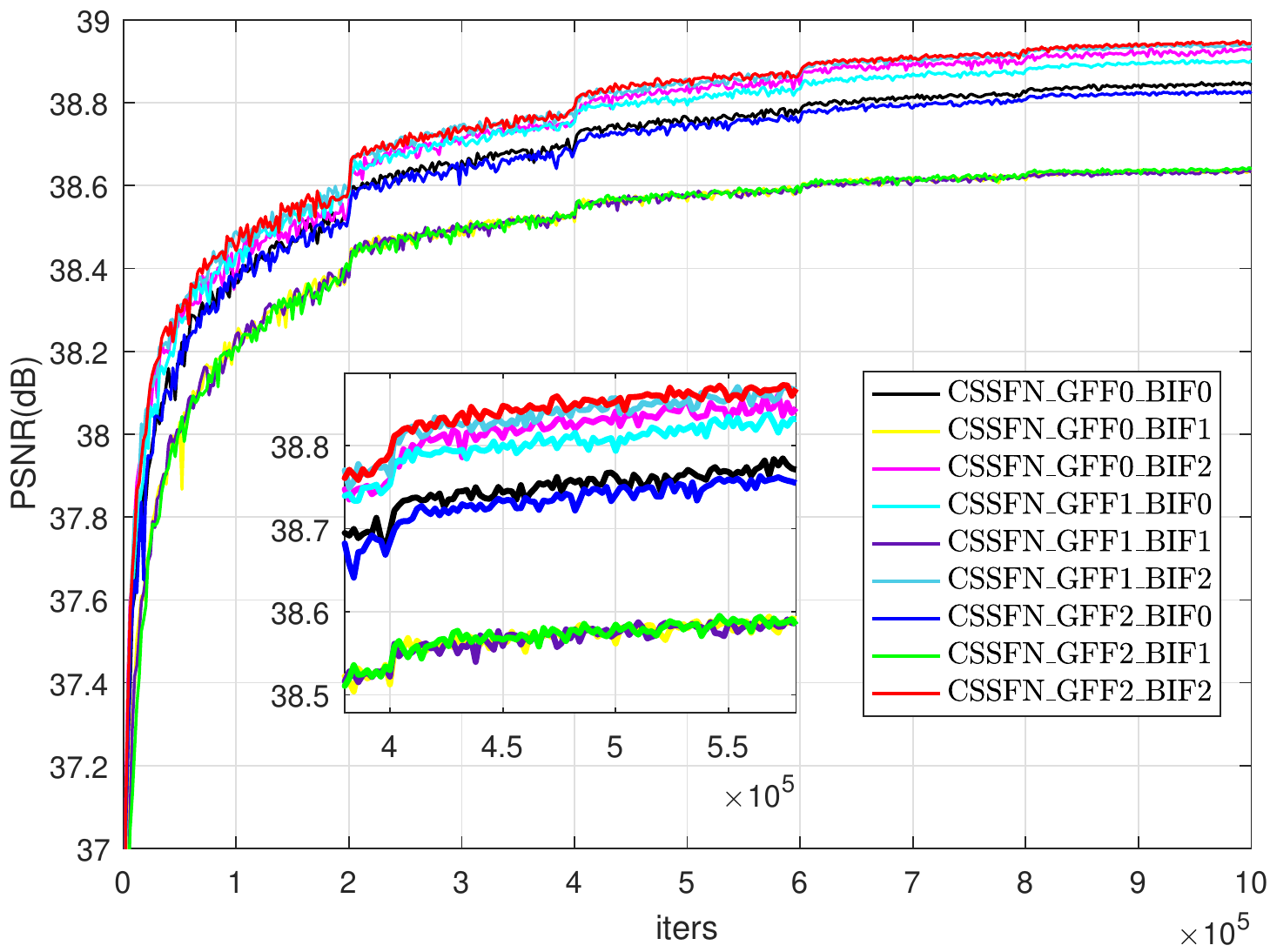}   \\
  \vspace{-4mm}
  \caption{Comparison of validation performance between different combinations of GFF and BIF. The PSNR curves are evaluated on $\mathcal{V}$(T2, BD) with $r = 2$ and correspond to the testing results in Table \ref{tab:2}.}
  \label{fig:plot_valid_compare}
\end{figure}

\begin{table*}[t]
  \centering
  \caption{The impact of the output width of serial fusion on the performance of the model. All the models are trained on $\mathcal{D}$(PD, BD) for one million iterations and tested on $\mathcal{T}$(PD, BD). The basic configuration is $c = 256$ and $m = n = 4$ (PSNR (dB) $|$ SSIM $|$ $P$ $|$ $D$).}
  \vspace{-2mm}
  \begin{tabular}{C{1.0cm}|C{0.4cm}||C{3.5cm}|C{3.5cm}|C{3.5cm}|C{3.5cm}}
    \toprule
       Width        & $r$       & $q = 2$                 & $q = 4$                 & $q = 8$                 & $q = 16$ \\
    \hhline{--||----} 
    \multirow{3}{*}{$c_o = \dfrac{c}{q}$} & $\times$2 & 41.45 $|$ 0.9898 $|$ 16.40M $|$ 59 & 41.30 $|$ 0.9896 $|$ 11.09M $|$ 91 & 41.20 $|$ 0.9893 $|$ 7.990M $|$ 155  & 40.99 $|$ 0.9889 $|$ 6.341M $|$ 283 \\
    & $\times$3 & 36.15 $|$ 0.9711 $|$ 19.35M $|$ 59 & 35.99 $|$ 0.9702 $|$ 14.04M $|$ 91 & 35.83 $|$ 0.9690 $|$ 10.95M $|$ 155 & 35.56 $|$ 0.9673 $|$ 9.294M $|$ 283 \\
    & $\times$4 & 33.71 $|$ 0.9520 $|$ 18.76M $|$ 60 & 33.60 $|$ 0.9509 $|$ 13.45M $|$ 92 & 33.38 $|$ 0.9484 $|$ 10.35M $|$ 156& 33.07 $|$ 0.9447 $|$ 8.703M $|$ 284 \\
    \hhline{--||----}
    \multirow{3}{*}{$c_o = 64$}  & $\times$2 & 41.32 $|$ 0.9895 $|$ 9.911M $|$ 59 & 41.33 $|$ 0.9896 $|$ 11.09M $|$ 91 & 41.35 $|$ 0.9896 $|$ 13.46M $|$ 155 & 41.28 $|$ 0.9895 $|$ 18.18M $|$ 283 \\
    & $\times$3 & 36.06 $|$ 0.9706 $|$ 12.86M $|$ 59 & 36.01 $|$ 0.9703 $|$ 14.04M $|$ 91 & 36.02 $|$ 0.9704 $|$ 16.41M $|$ 155 & 35.99 $|$ 0.9701 $|$ 21.13M $|$ 283 \\
    & $\times$4 & 33.57 $|$ 0.9506 $|$ 12.27M $|$ 60 & 33.59 $|$ 0.9509 $|$ 13.45M $|$ 92 & 33.59 $|$ 0.9506 $|$ 15.82M $|$ 156 & 33.59 $|$ 0.9508 $|$ 20.54M $|$ 284 \\
    \bottomrule
  \end{tabular}
  \label{tab:3}
\end{table*}

\begin{table*}[t]
  \centering
  \caption{Quantitative comparison between different algorithms on $\mathcal{T}$ (:, BD). The maximal PSNR (dB) and SSIM values of each row are marked in {\textcolor[rgb]{1,0,0}{red}}, and the second ones are marked in {\textcolor[rgb]{0,0,1}{blue}}.}
  \vspace{-2mm}
  \begin{tabular}{C{0.8cm}|C{0.4cm}||C{1.55cm}|C{1.55cm}||C{1.55cm}|C{1.55cm}|C{1.55cm}|C{1.55cm}|C{1.55cm}|C{1.55cm}}
    \toprule
     MR    & \multirow{2}{*}{$r$} & Bicubic & NLM & SRCNN  & VDSR  & RDN  & CSN  & CSSFN &  CSSFN \\
     images & & (2D) & \cite{Manjon2010Nonlocal} & \cite{Dong2016Image} & \cite{Kim2016Accurate} & \cite{Zhang2018Residual} & \cite{Zhao2018Channel} & ($q=4$) & ($q=2$) \\
    \hhline{--||--||------} 
    \multirow{3}{*}{PD}& $\times$2 & 35.04/0.9664 & 37.26/0.9773 & 38.96/0.9836 & 39.97/0.9861 & {40.31/0.9870} & {41.28/0.9895} & {\textcolor[rgb]{0,0,1}{41.30/0.9896}} & {\textcolor[rgb]{1,0,0}{41.45/0.9898}} \\
    & $\times$3 & 31.20/0.9230 & 32.81/0.9436 & 33.60/0.9516 & 34.66/0.9599 & {35.08/0.9628} & {35.87/0.9693} & {\textcolor[rgb]{0,0,1}{35.99/0.9702}} & {\textcolor[rgb]{1,0,0}{36.15/0.9711}}  \\
    & $\times$4 & 29.13/0.8799 & 30.27/0.9044 & 31.10/0.9181 & 32.09/0.9311 & {32.73/0.9387} & {33.40/0.9486} & {\textcolor[rgb]{0,0,1}{33.60/0.9509}} & {\textcolor[rgb]{1,0,0}{33.71/0.9520}} \\
    \hhline{--||--||------}
    \multirow{3}{*}{T1}& $\times$2 & 33.80/0.9525 & 35.80/0.9685 & 37.12/0.9761 & 37.67/0.9783 & {37.95/0.9795} & {38.27/0.9810} & {\textcolor[rgb]{0,0,1}{38.33/0.9812}} & {\textcolor[rgb]{1,0,0}{38.36/0.9813}} \\
                       & $\times$3 & 30.15/0.8900 & 31.74/0.9216 & 32.17/0.9276 & 32.91/0.9378 & {33.31/0.9430} & {33.53/0.9464} & {\textcolor[rgb]{0,0,1}{33.56/0.9468}} & {\textcolor[rgb]{1,0,0}{33.59/0.9471}} \\
                       & $\times$4 & 28.28/0.8312 & 29.31/0.8655 & 29.90/0.8796 & 30.57/0.8932 & {31.05/0.9042} & {31.23/0.9093} & {\textcolor[rgb]{0,0,1}{31.34/0.9102}} & {\textcolor[rgb]{1,0,0}{31.37/0.9104}} \\
    \hhline{--||--||------}
    \multirow{3}{*}{T2}& $\times$2 & 33.44/0.9589 & 35.58/0.9722 & 37.32/0.9796 & 38.65/0.9836 & {38.75/0.9838} & {39.71/0.9863} & {\textcolor[rgb]{0,0,1}{40.05/0.9870}} & {\textcolor[rgb]{1,0,0}{40.10/0.9871}} \\
                       & $\times$3 & 29.80/0.9093 & 31.28/0.9330 & 32.20/0.9440 & 33.47/0.9559 & {33.91/0.9591} & {34.64/0.9647} & {\textcolor[rgb]{0,0,1}{34.84/0.9661}} & {\textcolor[rgb]{1,0,0}{34.96/0.9668}} \\
                       & $\times$4 & 27.86/0.8611 & 28.85/0.8875 & 29.69/0.9052 & 30.79/0.9240 & {31.45/0.9324} & {32.05/0.9413} & {\textcolor[rgb]{0,0,1}{32.27/0.9441}} & {\textcolor[rgb]{1,0,0}{32.38/0.9453}} \\
    \bottomrule
  \end{tabular}
  \label{tab:4}
\end{table*}

\begin{table*}[t]
  \centering
  \caption{Quantitative comparison between different algorithms on $\mathcal{T}$ (:, TD). The maximal PSNR (dB) and SSIM values of each row are marked in {\textcolor[rgb]{1,0,0}{red}}, and the second ones are marked in {\textcolor[rgb]{0,0,1}{blue}}.}
  \vspace{-2mm}
  \begin{tabular}{C{0.8cm}|C{0.4cm}||C{1.55cm}|C{1.55cm}||C{1.55cm}|C{1.55cm}|C{1.55cm}|C{1.55cm}|C{1.55cm}|C{1.55cm}}
    \toprule
     MR     & \multirow{2}{*}{$r$} & Bicubic & NLM & SRCNN  & VDSR  & RDN  & CSN  & CSSFN &  CSSFN \\
     images & & (2D) & \cite{Manjon2010Nonlocal} & \cite{Dong2016Image} & \cite{Kim2016Accurate} & \cite{Zhang2018Residual} & \cite{Zhao2018Channel} & ($q=4$) & ($q=2$) \\
    \hhline{--||--||------}
    \multirow{3}{*}{PD}  & $\times$2 & 34.65/0.9625 & 36.18/0.9707 & 38.23/0.9802 & 39.89/0.9850 & 40.39/0.9862 & {41.77/0.9897} & {\textcolor[rgb]{0,0,1}{41.91/0.9900}} & {\textcolor[rgb]{1,0,0}{41.97/0.9902}} \\
    & $\times$3 & 30.88/0.9167 & 32.02/0.9324 & 32.90/0.9432 & 34.27/0.9555 & 35.00/0.9609 & {36.09/0.9697} & {\textcolor[rgb]{0,0,1}{36.23/0.9706}} & {\textcolor[rgb]{1,0,0}{36.32/0.9713}} \\
    & $\times$4 & 28.82/0.8713 & 29.27/0.8906 & 30.52/0.9078 & 31.69/0.9244 & 32.64/0.9362 & {33.51/0.9489} & {\textcolor[rgb]{0,0,1}{33.64/0.9501}} & {\textcolor[rgb]{1,0,0}{33.75/0.9514}} \\
    \hhline{--||--||------}
    \multirow{3}{*}{T1}  & $\times$2 & 33.38/0.9460 & 34.71/0.9581 & 36.52/0.9705 & 37.58/0.9760 & 38.08/0.9784 & {38.62/0.9813} & {\textcolor[rgb]{0,0,1}{38.67/0.9815}} & {\textcolor[rgb]{1,0,0}{38.76/0.9818}} \\
    & $\times$3 & 29.79/0.8793 & 30.83/0.9027 & 31.72/0.9187 & 32.57/0.9304 & 33.33/0.9416 & {33.68/0.9464} & {\textcolor[rgb]{0,0,1}{33.73/0.9469}} & {\textcolor[rgb]{1,0,0}{33.75/0.9472}} \\
    & $\times$4 & 27.96/0.8182 & 28.68/0.8439 & 29.31/0.8616 & 30.14/0.8818 & 31.00/0.9018 & {31.27/0.9092} & {\textcolor[rgb]{0,0,1}{31.35/0.9095}} & {\textcolor[rgb]{1,0,0}{31.39/0.9098}} \\
    \hhline{--||--||------}
    \multirow{3}{*}{T2}  & $\times$2 & 33.06/0.9541 & 34.56/0.9641 & 37.04/0.9773 & 38.74/0.9823 & 40.02/0.9826 & {40.47/0.9868} & {\textcolor[rgb]{0,0,1}{40.64/0.9872}} & {\textcolor[rgb]{1,0,0}{40.73/0.9874}} \\
    & $\times$3 & 29.50/0.9016 & 30.57/0.9197 & 31.80/0.9381 & 33.23/0.9515 & 33.99/0.9576 & {34.95/0.9653} & {\textcolor[rgb]{0,0,1}{35.12/0.9663}} & {\textcolor[rgb]{1,0,0}{35.23/0.9671}} \\
    & $\times$4 & 27.60/0.8511 & 28.37/0.8718 & 29.32/0.8960 & 30.51/0.9162 & 31.49/0.9301 & {32.28/0.9421} & {\textcolor[rgb]{0,0,1}{32.46/0.9440}} & {\textcolor[rgb]{1,0,0}{32.57/0.9453}} \\
    \bottomrule
  \end{tabular}
  \label{tab:5}
\end{table*}

\subsection{Channel Splitting and Serial Fusion Unit (CSSFU)}
\label{subsec:dcsu}
In the CSN network \cite{Zhao2018Channel}, the feature map transmitted to a channel splitting block (CSB) is first split into two branches with different network structures, which are then integrated together with the merge-and-run (MAR) mapping \cite{Hu2018Single,Zhao2017Deep}. In our CSSFN, the hierarchical feature is also split into several subfeatures with fewer channels. However, we do not transmit the information in a multi-branch way. Instead, the subfeatures are reintegrated into a single branch step-by-step through conv and concatenation operations, which we term as partially dense learning with channel splitting.

The input map of the $j$-th CSSFU in the $i$-th CSSFB is first split into $q$ subfeatures equally, i.e., $\{\mathbf{x}_{i-1,j-1}^{0},\ldots,\mathbf{x}_{i-1,j-1}^{q-1}\}$. Denote $\mathbf{z}_{i-1,j-1}^{k}$ as the output of the $k$-th 3$\times$3 conv layer in Fig.\ref{fig3}(b) (cubes in dark gray), which is followed by a ReLU layer \cite{Nair2010Rectified}. Then we have:
\begin{equation}
\label{CONV-DCSU}
  \mathbf{z}_{i-1,j-1}^{k} = \max\Big(0, \mathcal{H}_{3}^{c/q}([\mathbf{x}_{i-1,j-1}^{k-1}, \mathbf{z}_{i-1,j-1}^{k-1}])\Big),
\end{equation}
where $k = 1,2,\ldots,q$ and $\mathbf{z}_{i-1,j-1}^{0} = 0$. Therefore, all these subfeatures are reintegrated together and the network is in a single branch. Finally, we extend the channel of the last output feature, $\mathbf{z}_{i-1,j-1}^{q}$, by a 3$\times$3 channel extension layer at the end of the CSSFU:
\begin{equation}
\label{CHANN-EXT}
  \mathbf{x}_{i-1,j} = \mathcal{H}_{3}^{c}(\mathbf{z}_{i-1,j-1}^{q}),
\end{equation}
where $\mathbf{x}_{i-1,j}$ is the output feature map of the $j$-th CSSFU in the $i$-th CSSFB. It is worth pointing out that the purpose of channel splitting in this paper is not to form a multi-branch structure, but to be a preprocessing of the subsequent serial fusion. The single-branch structure makes the network deeper and narrower, which makes model training more unstable. This is part of the reason why we adopt DGFF to fuse inter-block features. Therefore, channel splitting and serial fusion can be regarded as ``stretching'' a relatively shallow but wider network into a deeper but narrower network.

\subsection{Network Depth and Model Parameters}
The network depth is usually defined as the length of the longest path from the input to the output \cite{Zhao2018Channel,Hu2018Single}. According to the entire structure of the proposed model, the depth of our CSSFN network is given by:
\begin{equation}
\label{DEPTH}
  D = n[1 + m \times (q + 1)] + s + 6,
\end{equation}
where $s$ denotes the depth of the upscale modula and depends on the specific value of the scaling factor $r$. Specifically, $s = 1$ for $r = 2$ or $r = 3$, and $s = 2$ for $r = 4$. The first ``1'' in (\ref{DEPTH}) corresponds to the compression layer at the beginning of each CSSFB, and the second one denotes the extension layer at the end of each CSSFU.

Table \ref{tab:1} collects the network depth ($D$) and the number of model parameters ($P$) of the proposed CSSFN model under several configuration conditions, where pseudo 3D execution implies that the model regards 96 slices of a 3D MR volume as 96 channels of a 2D image. As can be seen, all models need to determine about 10M $\sim$ 20M model parameters. The most similar model to our CSSFN is the CSN \cite{Zhao2018Channel}, so we display the comparison of network configuration between the CSN and our CSSFN in Fig.\ref{fig:param-compare}. We can observe that the proposed CSSFN model increases in network depth for both $q = 4$ and $q = 2$. However, it has fewer model parameters when $q = 4$ and more model parameters when $q = 2$.

\begin{figure*}[t]
  \centering
  \includegraphics[width=\textwidth]{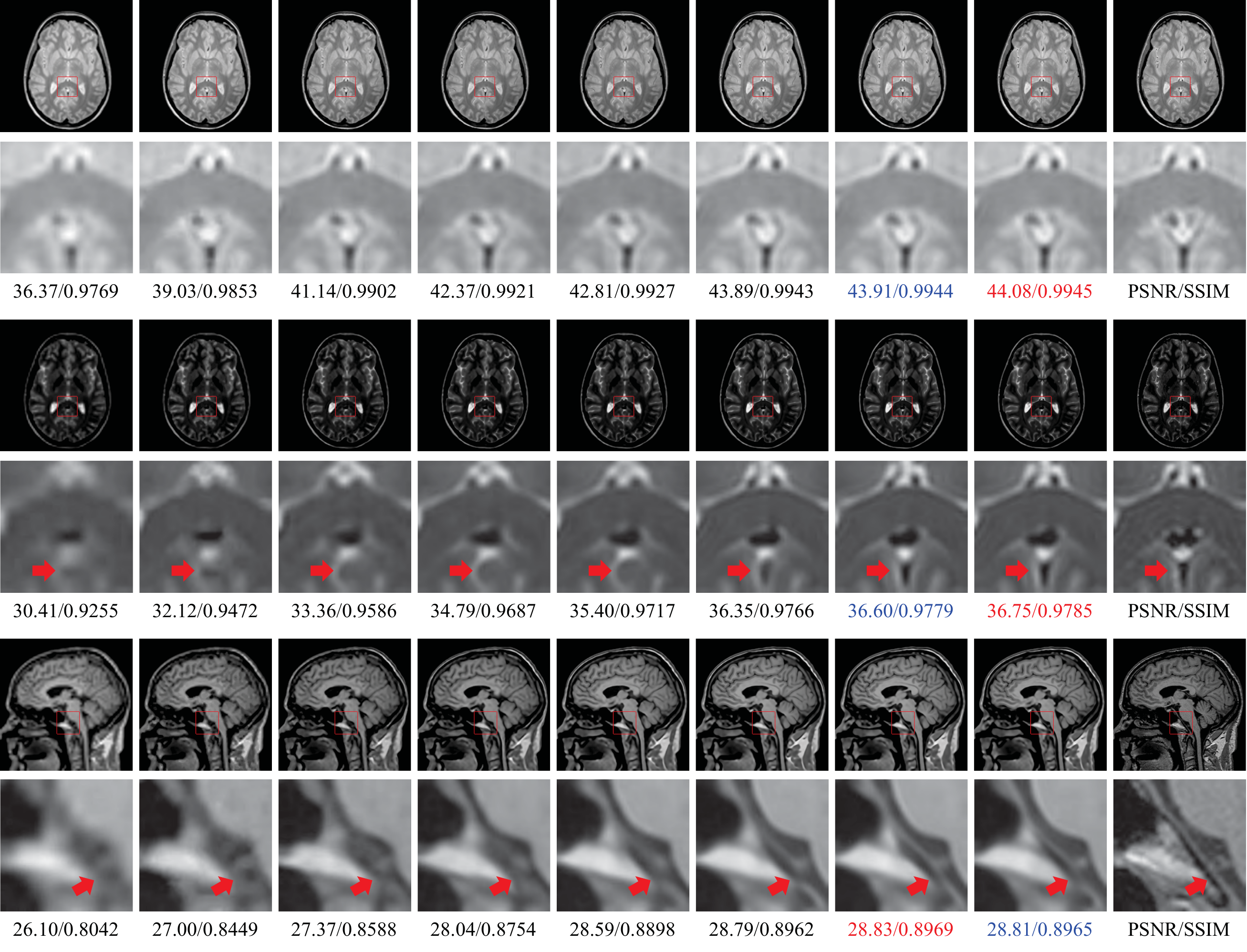}\\
  \vspace{-2mm}
  \caption{The visual comparison between several advanced SISR methods on a PD image with $r = 2$ (top), a T2 image with $r = 3$ (middle) and a T1 image with $r = 4$ (bottom). The image degradation is \textbf{bicubic degradation}. The first to the last columns are Bicubic, NLM \cite{Manjon2010Nonlocal}, SRCNN \cite{Dong2016Image}, VDSR \cite{Kim2016Accurate}, RDN \cite{Zhang2018Residual}, CSN \cite{Zhao2018Channel}, CSSFN-B4, CSSFN-B2 and the ground truth. The maximal PSNR (dB) and SSIM for each row are in red and the second ones are in blue.}
  \label{fig:vis_comp_bicubic_mix}
\end{figure*}

\begin{figure*}[t]
  \centering
  \includegraphics[width=\textwidth]{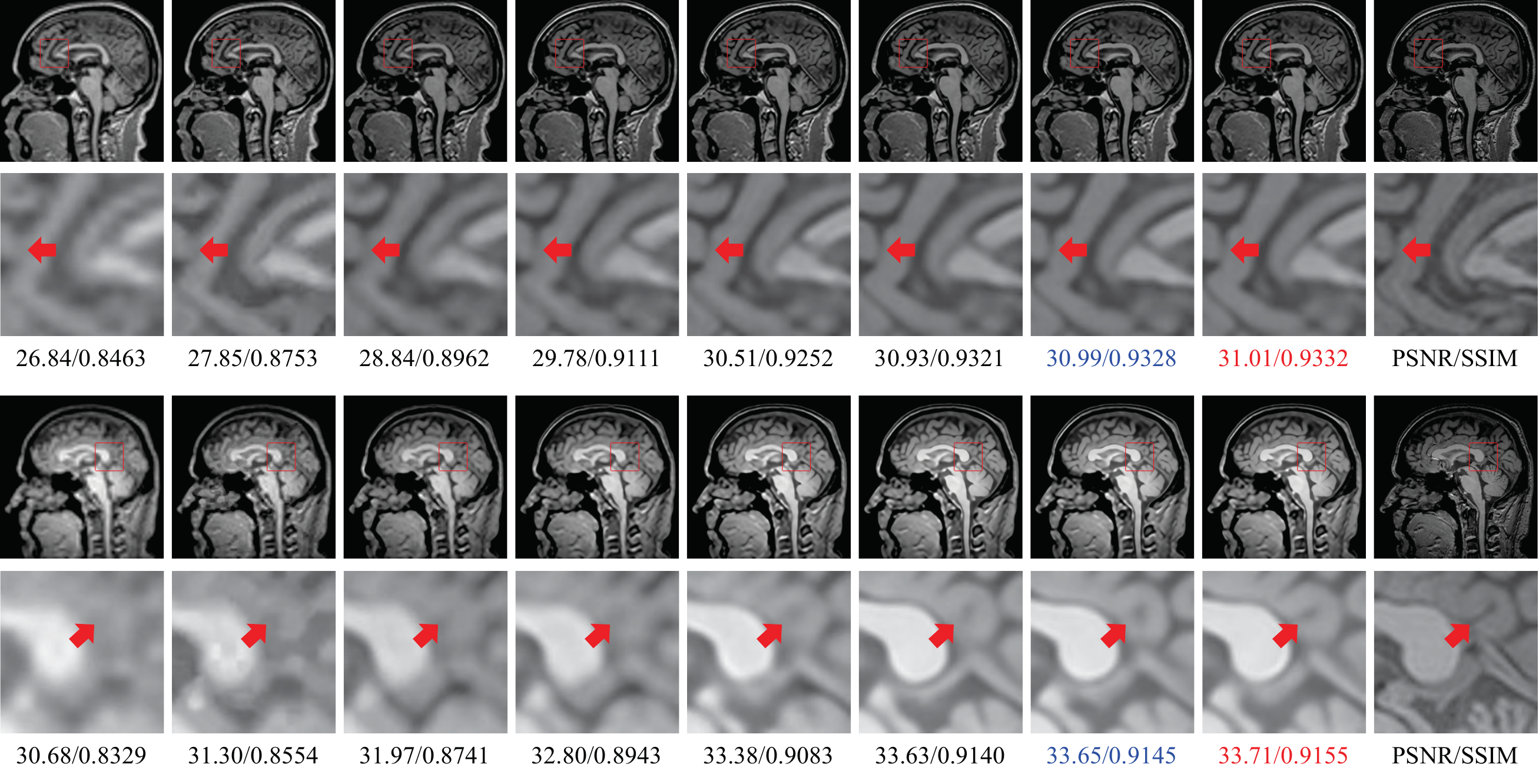}\\
  \vspace{-2mm}
  \caption{The visual comparison between several advanced SISR methods (\textbf{truncation degradation}) on T1 images with $r = 3$ (top) and $r = 4$ (bottom). The first to the last columns are Bicubic, NLM \cite{Manjon2010Nonlocal}, SRCNN \cite{Dong2016Image}, VDSR \cite{Kim2016Accurate}, RDN \cite{Zhang2018Residual}, CSN \cite{Zhao2018Channel}, CSSFN-B4, CSSFN-B2 and the ground truth. The maximal PSNR (dB) and SSIM for each row are marked in red, and the second ones are in blue.}
  \label{fig:vis_comp_truncation_T1}
\end{figure*}

\begin{figure*}[t]
  \centering
  \includegraphics[width=\textwidth]{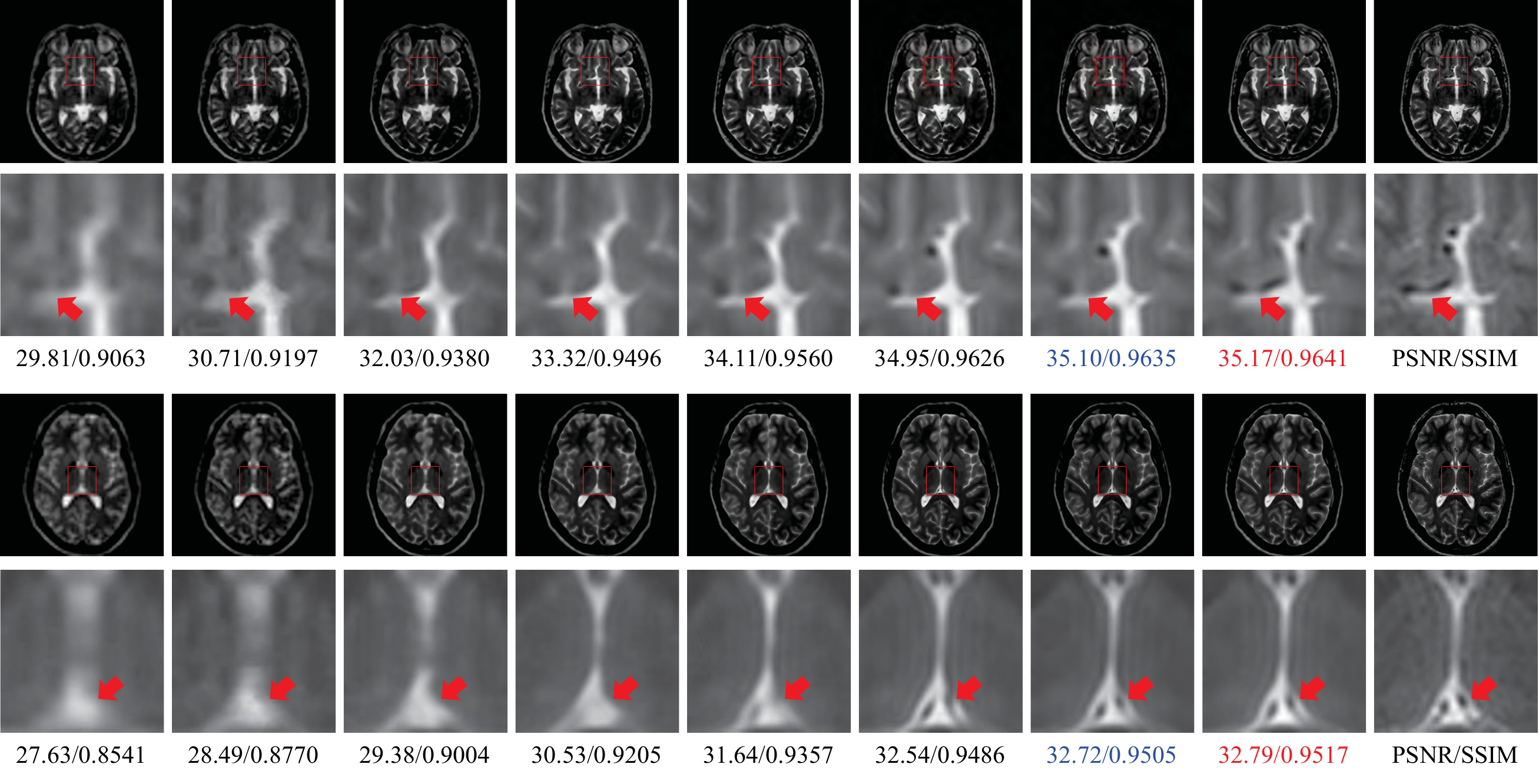}\\
  \vspace{-2mm}
  \caption{The visual comparison between several advanced SISR methods (\textbf{truncation degradation}) on T2 images with $r = 3$ (top) and $r = 4$ (bottom). The first to the last columns are Bicubic, NLM \cite{Manjon2010Nonlocal}, SRCNN \cite{Dong2016Image}, VDSR \cite{Kim2016Accurate}, RDN \cite{Zhang2018Residual}, CSN \cite{Zhao2018Channel}, CSSFN-B4, CSSFN-B2 and the ground truth. The maximal PSNR (dB) and SSIM for each row are marked in red, and the second ones are in blue.}
  \label{fig:vis_comp_truncation_T2}
\end{figure*}

\section{Experimental Results}
\label{sec:experiments}
In this section, we first briefly introduce the dataset used in this work and the details of model implementation. Then we investigate and analyze the structure of our network, including the influence of the way of global feature fusion (GFF) and branch information fusion (BIF), and the number of branches ($q$) on the model performance. Finally, the proposed method is compared with other advanced methods quantitatively and qualitatively. The frequently used peaks signal to noise ratio (PSNR) and structural similarity index metric (SSIM) \cite{Wang2004Image} are taken as the metrics of quantitative evaluation.

\subsection{Dataset and Implementation Details}
We utilize the same datasets as in \cite{Zhao2018Channel} to perform relevant experiments. They are derived from the IXI dataset and contain three structural MR volumes: T1-weighted, T2-weighted and PD-weighted images. Two LR image degradation models, i.e., bicubic downsampling (BD) and $k$-space truncation (TD), are implemented to simulate LR images. For convenience, the sub dataset with a specific type of MR images and degradation is also denoted as dataset name (MR type, degradation), just like \cite{Zhao2018Channel}. It is worthy noting that the all 3D MR images are cut to the size of 240$\times$240$\times$96, where 96 represents the number of slices of a 3D volume. If the model takes a single slice of a 3D volume as input, we call it pure 2D execution; if the model regards 96 slices as 96 channels of a 2D input, we term it as pseudo 3D execution, as shown in Table \ref{tab:1}.

The overall configuration of the proposed network is shown Fig.\ref{fig2} and Fig.\ref{fig3} with $c = 256$ and $m = n = q = 4$. The size of minibatch is set to 16. The kernel size of each convolutional layer is marked in Fig.\ref{fig2} and Fig.\ref{fig3}. For each convolution layer in CSSFU, we keep the channel size of the output feature is the same as that of subfeature maps, i.e., $c/q$, except that the last channel extension layer has $c$ output channels. For fair comparison, we also train the model with LR image patches of size 24$\times$24 with the corresponding HR patches. These training patches are further augmented by random horizontal flips and 90${}^{\circ}$ rotations, as in \cite{Zhao2018Channel,Lim2017Enhanced,Zhang2018Residual,Zhang2018Image}. All models are implemented in TensorFlow 1.11.0 and trained on a NVIDIA GeForce GTX 1080 Ti GPU for one million iterations. We apply xavier's method \cite{Glorot2010Understanding} to initialize model parameters. The optimizer to minimize the $L_1$ loss is the Adam algorithm \cite{Kingma2014ADAM} with ${\beta}_{1} = 0.9$, ${\beta}_{2} = 0.999$ and $\epsilon = 10^{-8}$. Learning rate is initialized as 0.0001 for all layers and halved at every 200K iterations, i.e., piecewise constant decay.

\subsection{Feature Fusion}
In this section, we investigate the effects of GFF and BIF on the performance of the model. To this end, we designed several structures for ablation investigation. For GFF, we compare the DGFF (Fig.\ref{fig2}) and the method adopted by \cite{Zhao2018Channel,Zhang2018Residual}, which we term as concat global feature fusion (CGFF). On the other hand, we compare the proposed serial fusion (SF, Fig.\ref{fig3}(b) and Fig.\ref{fig4}(c)) and the MAR mapping \cite{Zhao2017Deep} (Fig.\ref{fig4}(b)) for BIF. Note that the latter can be regarded as a way of parallel fusion for multi-branch structure. In addition, we have also constructed a benchmark structure without either GFF or BIF, where the GFF is removed from the entire network and the part in the dotted box in Fig.\ref{fig4}(c) is replaced with the basic Conv + ReLU structure in Fig.\ref{fig4}(a). Table \ref{tab:2} shows the results of comparison evaluated on $\mathcal{T}$(T2, BD), for SR$\times$2. As can be seen, the benchmark structure without channel splitting (the 3rd column) can achieve 39.90dB PSNR, a relatively good result. This is probably because that the stage mapping in the benchmark structure (refer to Fig.\ref{fig4}(a) and Fig.\ref{fig4}(c)) evolves into the residual block structure of EDSR model \cite{Lim2017Enhanced}.

For convenience, we use two numbers to represent different GFF and BIF methods, 0 for the benchmark structure, GFF1 for CGFF, GFF2 for DGFF, BIF1 for MAR (Fig.\ref{fig4}(b)), and BIF2 for SF (Fig.\ref{fig4}(c)). According to the 6th, 8th and 10th columns, we can observe that the MAR mapping degrades the model performance seriously when $q = 4$, which implies that one can not improve the SR performance by simply increasing the branch number of the CSN \cite{Zhao2018Channel}. On the contrary, our SF strategy can boost the performance of the model (the 7th, 9th and 11th columns), compared with the benchmark structure (the 3rd column). Another interesting thing is that CGFF performs significantly better than DGFF if without channel splitting (the 4th column vs. the 5th column). However, the situation is reversed if with channel splitting (the 8th column vs. the 10th column and the 9th column vs. the 11th column). This shows that the combination of DGFF and SF can better promote the flow of information in the network and improve the model performance.

We also visualize the convergence process of these models in Fig.\ref{fig:plot_valid_compare}. These validation curves are almost consistent with the results displayed in table \ref{tab:2} and the above analysis, and the comparison is more obvious. Both the quantitative results in Table \ref{tab:2} and the visualization of the validation process demonstrate the effectiveness and benefits of the proposed SF and the combination with DGFF.

\subsection{Channel Splitting}
The output width of serial fusion, i.e., $c/q$, will be changed according to the number of subfeatures in previous settings. In this case, the number of model parameters, $P$, will decrease as $q$ increases. However, if we set the output channel of serial fusion to a fixed value, then $P$ will increase with the increase of $q$. Denote the output channel of serial fusion as $c_o$, we study the effects of the number of subfeatures and $c_o$ on the model performance. To this end, we train the CSSFN model with different configurations with $\mathcal{D}$(PD, BD) and collect the results in Table \ref{tab:3}.

\subsubsection{Unfixed Output Width}
In this case, $c_o = c/q$. According to rows 2, 3 and 4 of Table \ref{tab:3}, we can see that the model performance degrades with the increase of $q$. This is easy to understand, because the increase of $q$ reduces the parameters of the model sharply, although it superficially increases the network depth.

\subsubsection{Fixed Output Width}
We set $c_o = 64$ for comparison in this case. As shown in rows 5, 6 and 7 of Table \ref{tab:3}, we can not obtain significant performance gains by increasing the number of subfeatures, $q$. This result is strange, because the parameters of the model ($P$) and the depth of the network ($D$) increase with the increase of $q$, but the model performance does not improve and even seems to decline (e.g., SR$\times$3). Increasing $q$, in fact, will make it more difficult to effectively integrate these subfeatures. Thus, channel discrimination should \textit{not} aggressively increase the number of subfeatures unless there exists a more effective information fusion mechanism.

From Table \ref{tab:3}, we can generally draw the conclusion that the performance of the model is mainly related to $c_o$ when the overall framework remains unchanged, regardless of how the model parameters and the network depth change with $q$.

\subsection{Comparison with Other Methods}
To illustrate the effectiveness and the superiority of the proposed CSSFN model, we compare it with several typical SISR methods in this section, including NLM \cite{Manjon2010Nonlocal}, SRCNN \cite{Dong2016Image}, VDSR \cite{Kim2016Accurate}, RDN \cite{Zhang2018Residual}, EDSR \cite{Lim2017Enhanced} and CSN \cite{Zhao2018Channel}. Among them, NLM is a classic conventional method for MR image upsampling, and SRCNN and VDSR are two typical lightweight CNN-based methods, while RDN and EDSR are two representative CNN-based methods that have large-scale model parameters. Some results are directly cited from \cite{Zhao2018Channel} due to the same datasets and image degradation, while others are obtained by retraining the corresponding models.

\subsubsection{Bicubic Degradation (BD)}
As one of the most common image degradation models for simulating LR images, bicubic degradation simply utilizes the bicubic kernel to reduce image size in image space. Table \ref{tab:4} exhibits the quantitative results of the compared methods over the testing datasets of PD, T1 and T2 MR images for SR$\times$2, SR$\times$3 and SR$\times$4. It can be seen that our CSSFN model surpasses the CSN model \cite{Zhao2018Channel} and achieves the best SR performance on all MR image types and all scaling factors. In particular, the CSSFN has relatively few model parameters when $q = 4$ but also gives excellent SR performance.

Fig.\ref{fig:vis_comp_bicubic_mix} shows the visual effect comparison between these SR methods on a PD (top), T2 (middle) and T1 (bottom) image with SR$\times$2, SR$\times$3 with SR$\times$4, respectively. From the first row of Fig.\ref{fig:vis_comp_bicubic_mix}, we can see that most of the CNN-based methods (e.g., VDSR \cite{Kim2016Accurate} and RDN \cite{Zhang2018Residual}) can surpass the traditional methods (bicubic and NLM \cite{Manjon2010Nonlocal}) by a large marge, achieving a perfect approximation to the ground truth. It is not easy to observe the differences between CNN-based methods when $r = 2$, but we can see that the CSSFN model is the most accurate one from the quantitative evaluation below each image. With the increase of $r$, the difference in the visual effect becomes more obvious. For instance, in the T2 image of the second row, the black hole indicated by the {\textcolor[rgb]{1,0,0}{red arrow}} has the most similar shape to the ground truth in the result of our method. The black hole, however, almost completely disappears in the results of some methods, e.g., bicubic, NLM \cite{Manjon2010Nonlocal} and even SRCNN \cite{Dong2016Image}, VDSR \cite{Kim2016Accurate} and RDN \cite{Zhang2018Residual}. Similar results can also be observed in the T1 image of the third row. In the ground truth, there is a continuous white ridge indicated by the {\textcolor[rgb]{1,0,0}{red arrow}}. However, it can hardly be identified in the results of NLM \cite{Manjon2010Nonlocal}, SRCNN \cite{Dong2016Image} and VDSR \cite{Kim2016Accurate}. In the results of RDN \cite{Zhang2018Residual} and CSN \cite{Zhao2018Channel}, the ridge is disconnected. Our CSSFN models present a obviously visible and continuous ridge that is closest to the ground truth.

\begin{table*}[t]
  \centering
  \caption{Quantitative comparison between several methods in case of pseudo 3D execution. The maximal PSNR (dB) and SSIM of each comparative group ($\mathcal{T}$(:, BD) and $\mathcal{T}$(:, TD)) are marked in {\textcolor[rgb]{1,0,0}{red}}, and the second ones are marked in {\textcolor[rgb]{0,0,1}{blue}}.}
  \vspace{-2mm}
  \begin{tabular}{C{0.8cm}|C{0.4cm}|C{1.55cm}|C{1.55cm}|C{1.55cm}|C{1.55cm}||C{1.55cm}|C{1.55cm}|C{1.55cm}|C{1.55cm}}
    \toprule
       & \multirow{3}{*}{$r$} &  \multicolumn{4}{c||}{Bicubic Downsampling Degradation $\mathcal{T}$(:, BD))}  &  \multicolumn{4}{c}{$k$-space Truncation Degradation $\mathcal{T}$(:, TD))}         \\
       \cmidrule{3-10}
       &                      & EDSR                   & CSN                    & CSSFN     & CSSFN     & EDSR  & CSN  & CSSFN &  CSSFN \\
       &                      & \cite{Lim2017Enhanced} & \cite{Zhao2018Channel} & ($q = 4$) & ($q = 2$) & \cite{Lim2017Enhanced} & \cite{Zhao2018Channel} & ($q=4$) & ($q=2$) \\
    \hhline{----------}
    \multirow{3}{*}{PD}  & $\times$2 & 39.87/0.9857 & 40.15/0.9865 & {\textcolor[rgb]{0,0,1}{40.28/0.9869}} & {\textcolor[rgb]{1,0,0}{40.34/0.9871}} & 39.47/0.9837 & 39.50/0.9839 & {\textcolor[rgb]{0,0,1}{39.80/0.9849}} & {\textcolor[rgb]{1,0,0}{39.91/0.9853}} \\
    & $\times$3 & 34.39/0.9578 & 34.68/0.9598 & {\textcolor[rgb]{1,0,0}{34.78}}/{\textcolor[rgb]{0,0,1}{0.9609}} & {\textcolor[rgb]{0,0,1}{34.76}}/{\textcolor[rgb]{1,0,0}{0.9611}} & 33.97/0.9531 & 34.12/0.9540 & {\textcolor[rgb]{1,0,0}{34.24/0.9554}} & {\textcolor[rgb]{0,0,1}{34.15/0.9550}} \\
    & $\times$4 & 31.80/0.9284 & {\textcolor[rgb]{0,0,1}{32.19}}/0.9325 & {\textcolor[rgb]{1,0,0}{32.21/0.9332}} & 32.11/{\textcolor[rgb]{0,0,1}{0.9329}} & 31.44/0.9219 & {\textcolor[rgb]{0,0,1}{31.72}}/0.9246 & {\textcolor[rgb]{1,0,0}{31.78}}/{\textcolor[rgb]{0,0,1}{0.9252}} & 31.68/{\textcolor[rgb]{1,0,0}{0.9257}} \\
    \hhline{----------}
    \multirow{3}{*}{T1}  & $\times$2 & 37.56/0.9774 & 37.60/0.9778 & {\textcolor[rgb]{0,0,1}{37.74/0.9786}} & {\textcolor[rgb]{1,0,0}{37.81/0.9789}} & 37.09/0.9741 & 36.99/0.9737 & {\textcolor[rgb]{0,0,1}{37.18/0.9748}} & {\textcolor[rgb]{1,0,0}{37.25/0.9754}} \\
    & $\times$3 & 32.76/0.9347 & 32.83/0.9360 & {\textcolor[rgb]{1,0,0}{32.86}}/{\textcolor[rgb]{0,0,1}{0.9362}} & {\textcolor[rgb]{0,0,1}{32.85}}/{\textcolor[rgb]{1,0,0}{0.9366}} & 32.27/0.9274 & 32.25/0.9266 & {\textcolor[rgb]{1,0,0}{32.34/0.9276}} & {\textcolor[rgb]{0,0,1}{32.32/0.9275}} \\
    & $\times$4 & 30.46/0.8902 & 30.53/0.8915 & {\textcolor[rgb]{0,0,1}{30.58/0.8919}} & {\textcolor[rgb]{1,0,0}{30.61/0.8923}} & 30.04/0.8803 & 30.07/0.8794 & {\textcolor[rgb]{0,0,1}{30.09/0.8795}} & {\textcolor[rgb]{1,0,0}{30.14/0.8812}} \\
    \hhline{----------}
    \multirow{3}{*}{T2}  & $\times$2 & 38.28/0.9824 & 38.53/0.9831 & {\textcolor[rgb]{0,0,1}{38.79/0.9836}} & {\textcolor[rgb]{1,0,0}{38.92/0.9842}} & 38.11/0.9803 & 38.20/0.9807 & {\textcolor[rgb]{0,0,1}{38.54/0.9817}} & {\textcolor[rgb]{1,0,0}{38.92/0.9842}} \\
    & $\times$3 & 33.15/0.9528 & 33.36/0.9547 & {\textcolor[rgb]{0,0,1}{33.46/0.9556}} & {\textcolor[rgb]{1,0,0}{33.50/0.9559}} & 32.89/0.9482 & 33.00/0.9490 & {\textcolor[rgb]{0,0,1}{33.21/0.9512}} & {\textcolor[rgb]{1,0,0}{33.26/0.9518}} \\
    & $\times$4 & 30.52/0.9198 & 30.81/0.9231 & {\textcolor[rgb]{1,0,0}{30.93/0.9242}} & {\textcolor[rgb]{0,0,1}{30.89/0.9241}} & 30.31/0.9137 & 30.54/0.9163 & {\textcolor[rgb]{1,0,0}{30.62/0.9182}} & {\textcolor[rgb]{0,0,1}{30.58/0.9178}} \\
    \bottomrule
  \end{tabular}
  \label{tab:6}
\end{table*}

\begin{figure*}[t]
  \centering
  \subfigure[$\mathcal{V}$(T1, BD) with $r = 2$]{\label{bicubic_T1_X2_C96_valid_compare}
  \begin{minipage}[t]{0.235\textwidth}
    \centering
    \includegraphics[scale = 0.28]{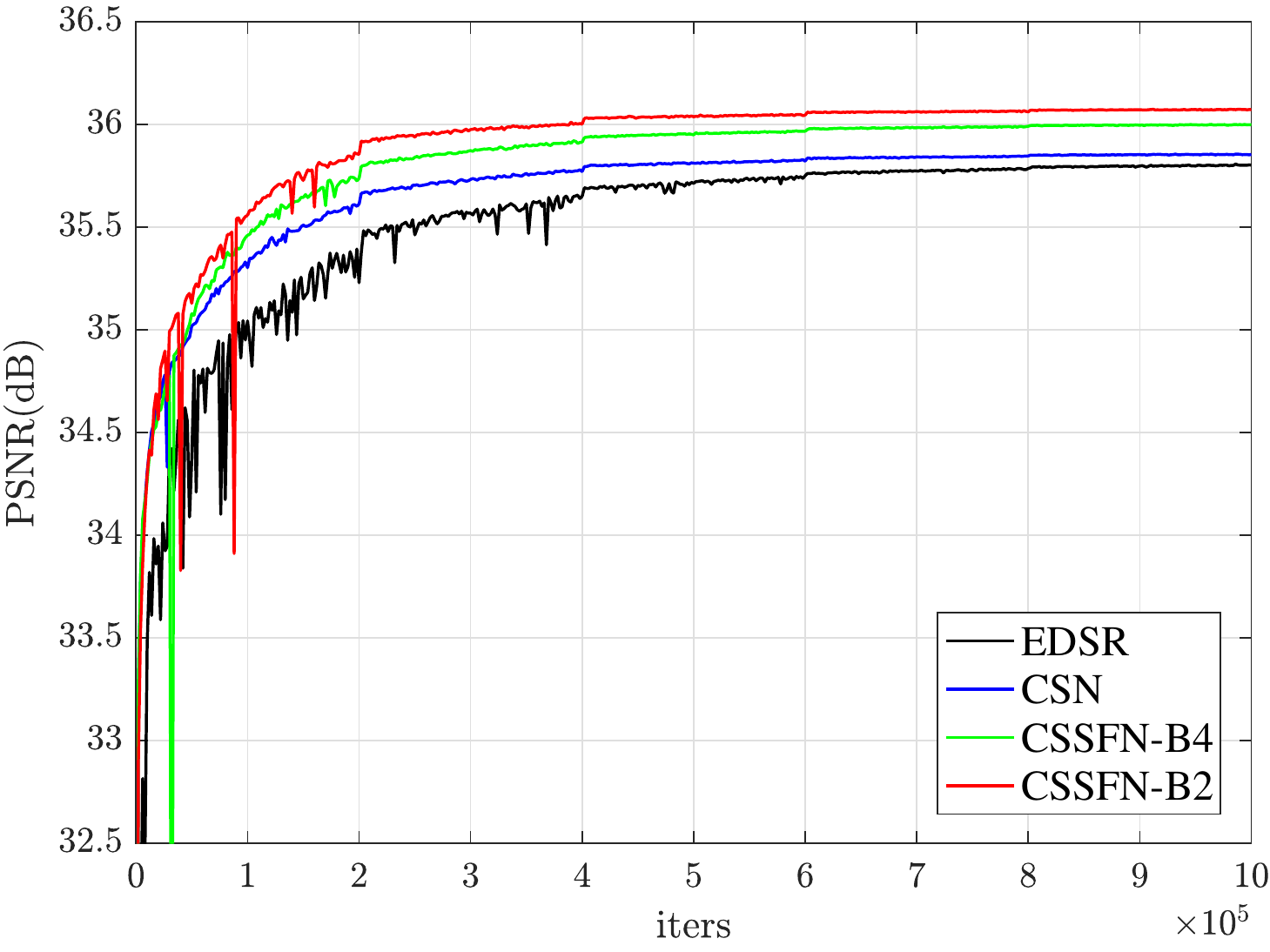}
  \end{minipage}}
  \subfigure[$\mathcal{V}$(T2, TD) with $r = 3$]{\label{truncation_T2_X3_C96_valid_compare}
  \begin{minipage}[t]{0.235\textwidth}
    \centering
    \includegraphics[scale = 0.28]{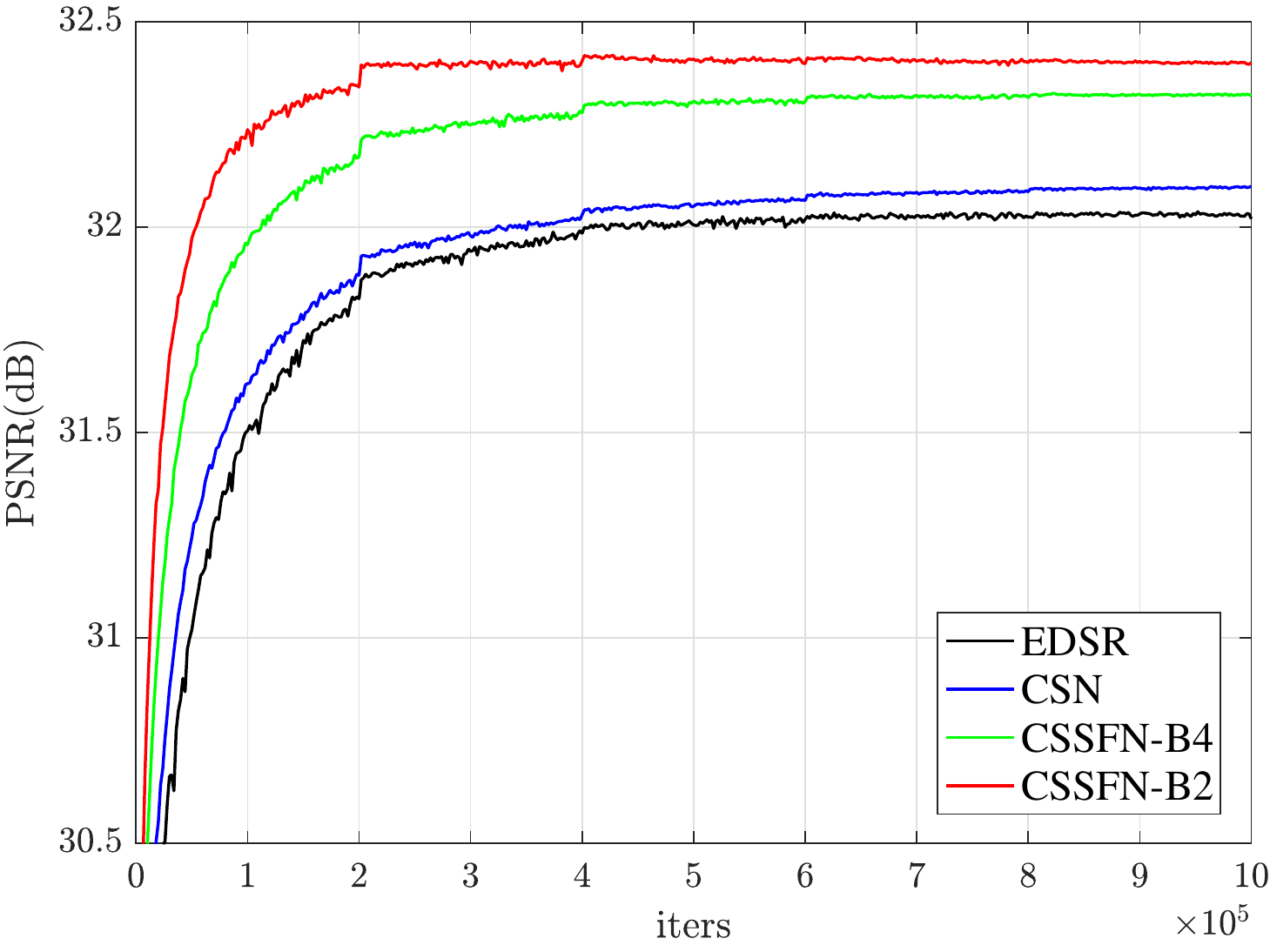}
  \end{minipage}}
  \subfigure[$\mathcal{V}$(T2, TD) with $r = 4$]{\label{truncation_T2_X4_C96_valid_compare}
  \begin{minipage}[t]{0.235\textwidth}
    \centering
    \includegraphics[scale = 0.28]{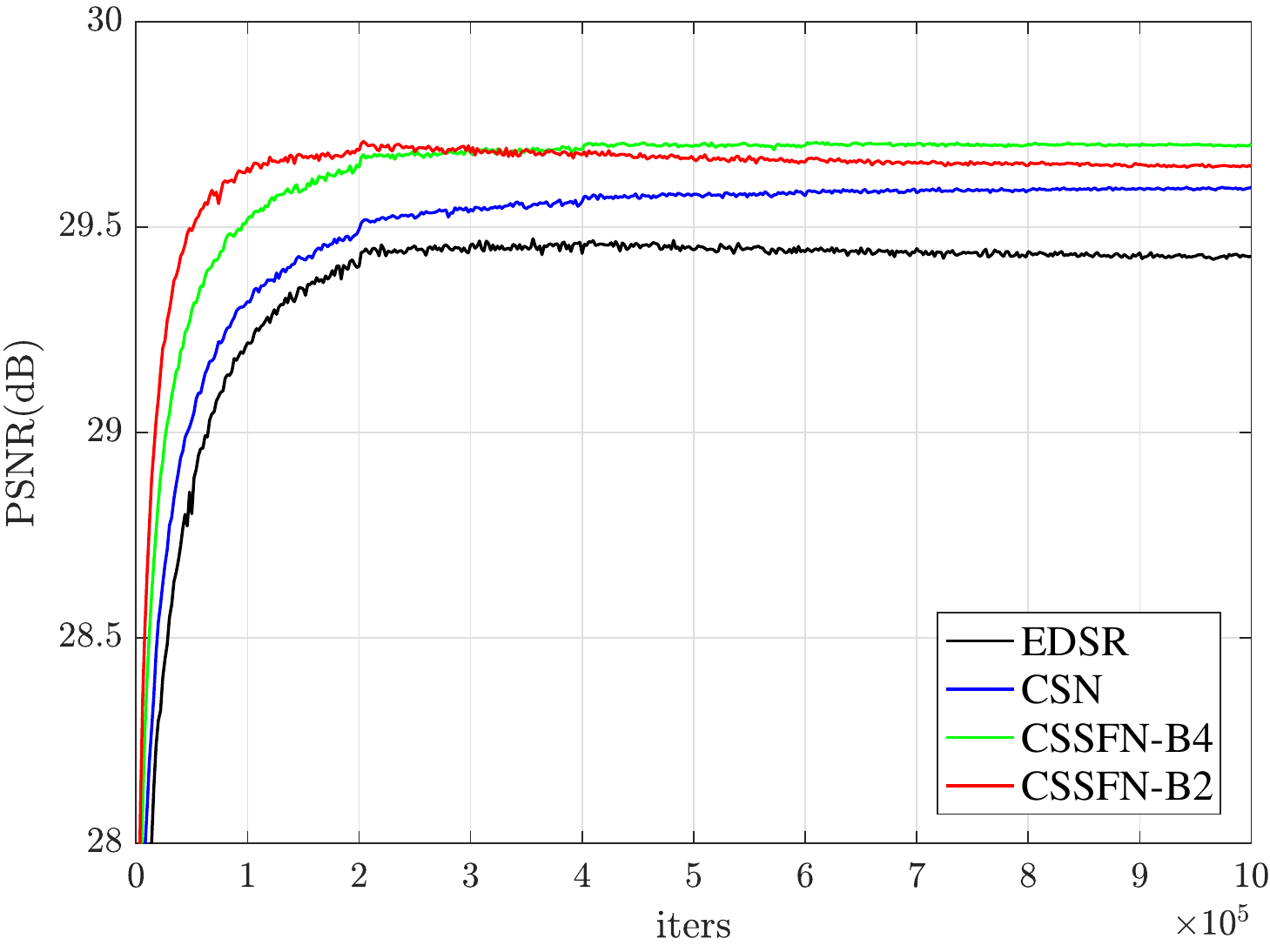}
  \end{minipage}}
  \subfigure[$\mathcal{V}$(PD, BD) with $r = 4$]{\label{bicubic_PD_X4_C96_valid_compare}
  \begin{minipage}[t]{0.235\textwidth}
    \centering
    \includegraphics[scale = 0.28]{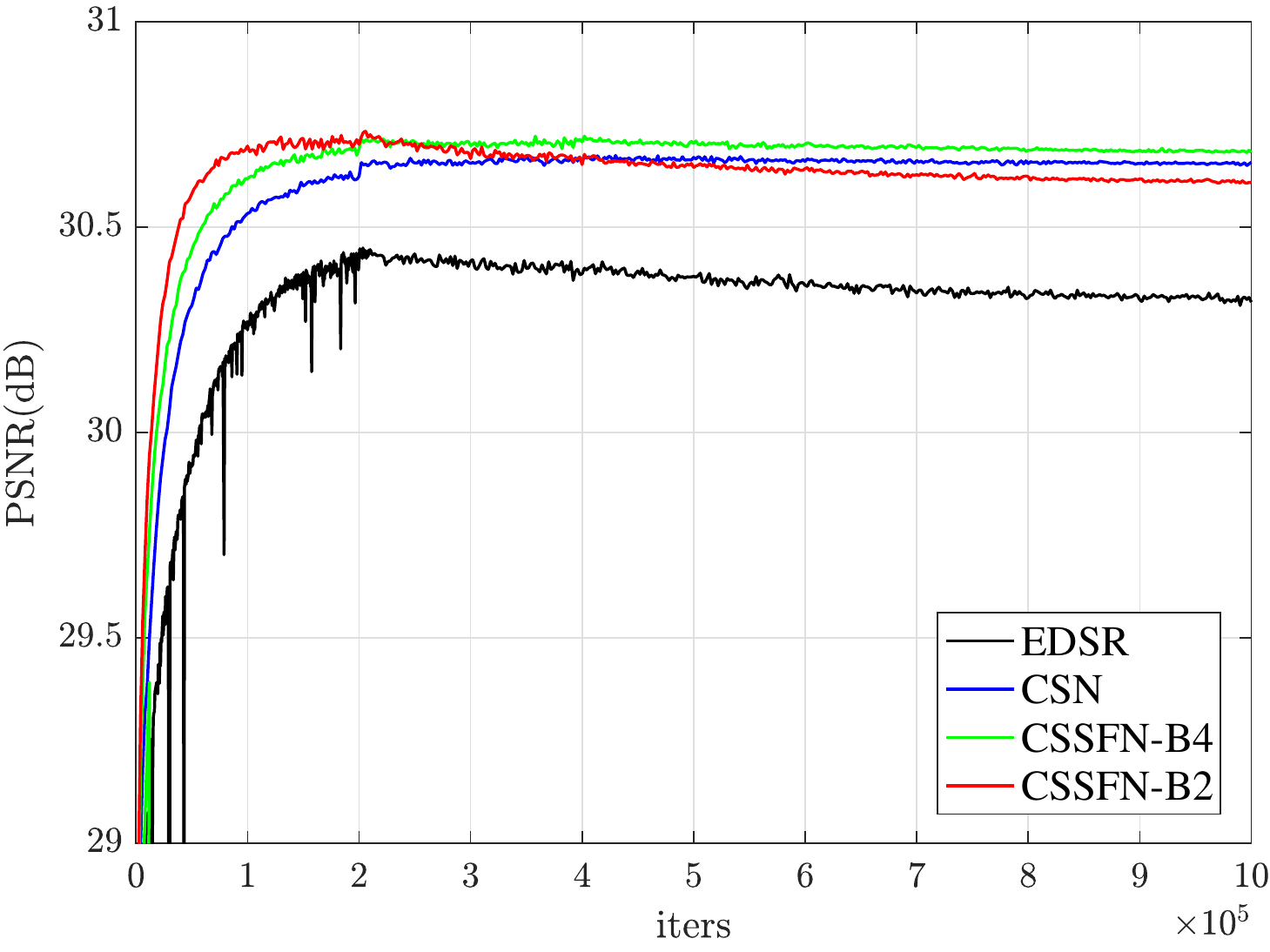}
  \end{minipage}}
  \vspace{-2mm}
  \caption{The validation performance comparison of the compared methods on several randomly selected sub datasets, in terms of pseudo 3D execution. It can be observed that the severity of model performance degradation due to over-fitting/under-fitting: (a) $<$ (b) $<$ (c) $<$ (d).}\label{visual_coompare_valid_C96}
\end{figure*}

\subsubsection{Truncation Degradation (TD)}
The $k$-space truncation of the HR image is a process that simulates the real MR image acquisition process where a LR image is scanned by reducing acquisition lines in phase and slice encoding directions \cite{Zhao2018Channel}. When the scaling factor $r$ is the same, the $k$-space truncation often reduces image quality more seriously than the bicubic downsampling due to the ``steep'' loss of the $k$-space data. This can be verified by the fact that bicubic interpolation performs better in bicubic downsampling than in $k$-space truncation (see the 3rd column of Table \ref{tab:4} and Table \ref{tab:5}). Table \ref{tab:5} gives the quantitative results of the compared methods in terms of the truncation degradation. Again, our CSSFN model exhibits the best SR performance on all MR image types and all scaling factors. But more importantly, the CSSFN models (and the CSN model \cite{Zhao2018Channel}) perform better than in the case of the bicubic downsampling, which implies that the proposed CSSFN model is more suitable for MR image super-resolution.

Visually, we can more easily observe the advantages of the proposed method to other methods. Fig.\ref{fig:vis_comp_truncation_T1} displays the visual effect of the compared methods on two T1 images with $r = 3$ and $r = 4$ respectively. The proposed method recovers images with clearer and sharper edges, thus making them more faithful to the ground truth. For example, there exists a dark line in the ground truth of the 2nd row of Fig.\ref{fig:vis_comp_truncation_T1} (indicated by the {\textcolor[rgb]{1,0,0}{red arrow}}), but only the CSSFN-B2 gives a credible hint of this dark line. Fig.\ref{fig:vis_comp_truncation_T2} shows the results on two T2 images with $r = 3$ and $r = 4$ respectively. In this case, the visual advantage of our CSSFN model is more observable. The quantitative results below each image also confirm the conclusion.

\subsubsection{Pseudo 3D Execution}
One of the major problem of training large-scale models with MR images is the degradation of training samples, e.g., it is difficult to successfully train the original EDSR model \cite{Lim2017Enhanced} with PD images \cite{Zhao2018Channel}. This issue can be alleviated by pseudo 3D execution at the cost of accuracy reduction. To compare our CSSFN model with other methods in this case, we also conduct the comparative experiments of pseudo 3D execution.

\begin{figure}[t]
  \centering
  \includegraphics[width=0.48\textwidth]{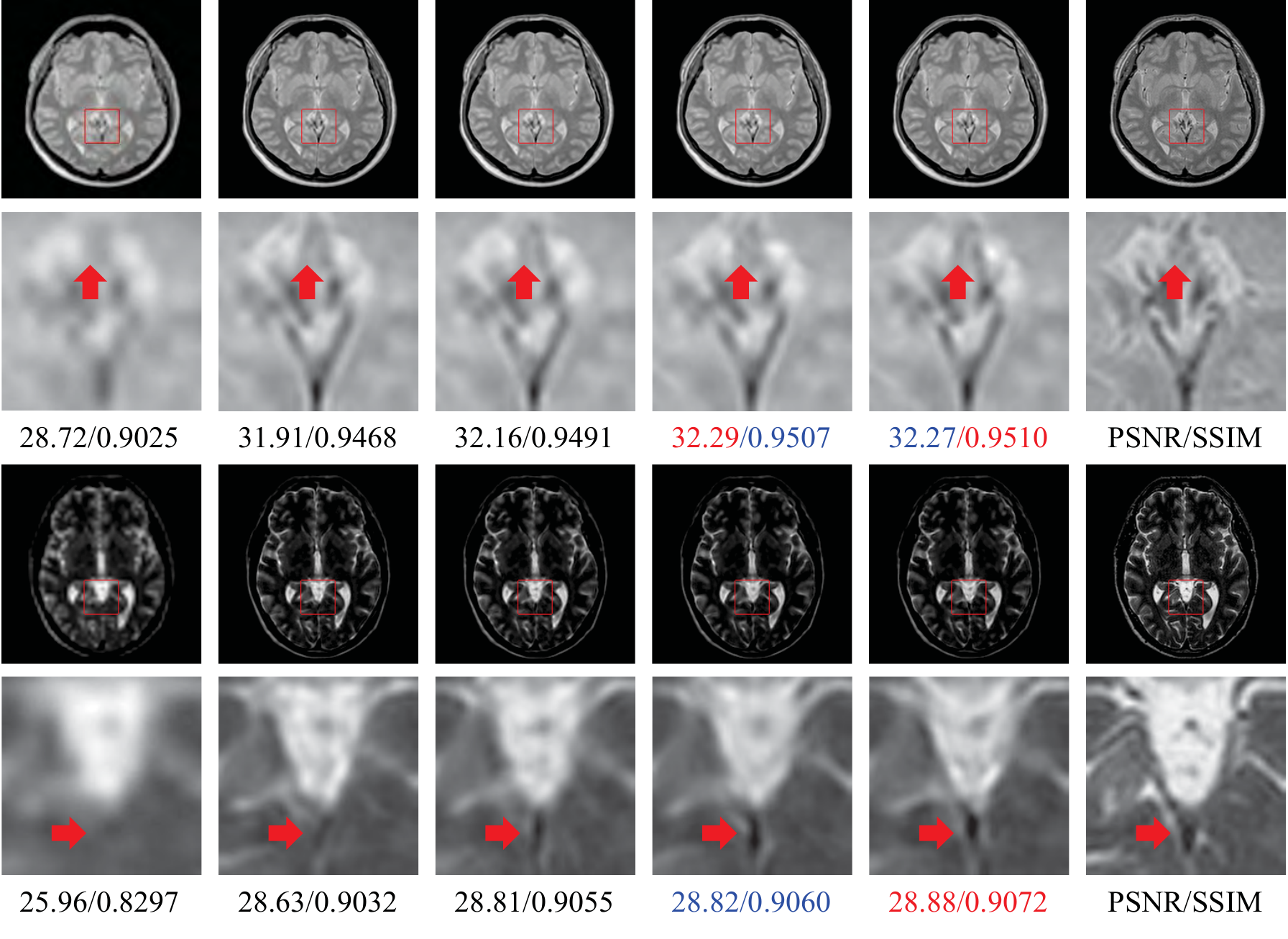}\\
  \vspace{-2mm}
  \caption{The visual comparison of pseudo 3D execution on a PD image with $r = 3$ and a T2 image with $r = 4$ (\textbf{bicubic degradation}). Left to right: Bicubic, EDSR \cite{Lim2017Enhanced}, CSN \cite{Zhao2018Channel}, CSSFN-B4, CSSFN-B2 and the ground truth.}
  \label{fig:vis_compare_C96}
\end{figure}

Table \ref{tab:6} shows the quantitative comparison between EDSR \cite{Lim2017Enhanced}, CSN \cite{Zhao2018Channel}, CSSFN-B4 and CSSFN-B2 in this case, with both image degradations. The performance improvement of the proposed CSSFN models is obvious. However, the advantage of CSSFN-B2 over CSSFN-B4 seems to be diminished when comparing with pure 2D execution, such as $\mathcal{T}$(PD, BD) with $r = 4$ and $\mathcal{T}$(T1, TD) with $r = 3$. In some cases, the PSNR performance of CSSFN-B2 is even worse than that of CSN \cite{Zhao2018Channel}, e.g., $\mathcal{T}$(PD, TD) with $r = 4$. Actually, this is primarily due to the underfitting or overfitting caused by training sample reduction, as shown in Fig.\ref{visual_coompare_valid_C96}. It can be seen that CSSFN-B2 essentially performs better than CSSFN-B4 in that it converges faster and has higher PSNR maximums. Besides, only CSN \cite{Zhao2018Channel} did not show obvious overfitting/underfitting in all cases. Fig.\ref{fig:vis_compare_C96} shows the visual comparison between these methods in terms of pseudo 3D execution, from which we can also observe that the proposed methods provide a clearer indication of the underlying structure, compared with other methods. Despite the overfitting/underfitting, we still train the models for one million iterations to keep the training iterations consistent with those of other comparative methods.

\section{Discussion}
\label{sec:discussion}
\subsection{Channel Discrimination Ability}
In the CSN model \cite{Zhao2018Channel}, the channel discrimination ability of the model is achieved by different branch structures. The hierarchical features is divided into 2 parts by channel splitting and fused together by the marge-and-run mapping \cite{Zhao2017Deep}. But the propagation paths of each subfeature have different branch structures. In the proposed CSSFN model, the subfeatures are also processed discriminatorily though they are placed in a single branch, in that the concat connections locate them at different depths of the network. This can be regarded as the fine-grained hierarchy of the intermediate features, thus realizing partial continuous memory mechanism \cite{Zhang2018Residual}, which is believed to be beneficial to the feedback propagation of the network \cite{He2015Identity}.

From Table \ref{tab:3}, we can observe that the model performance degrades gradually with the increase of subfeatures ($q$) when $c_o = c/q$. However, the increase of model parameters leads no obvious performance gains when $c_o = 64$. Thus, we speculate that the degradation of model performance is mainly because that increasing subfeatures will also increase the difficulty of information fusion. Further more, if more effective information fusion mechanisms are explored, the trade-off between model performance and model scale can be further improved.

\subsection{Image Degradation Model}
\begin{figure}[t]
  \centering
  \subfigure[]{\label{space_domain}
  \begin{minipage}[t]{0.23\textwidth}
    \centering
    \includegraphics[scale = 0.28]{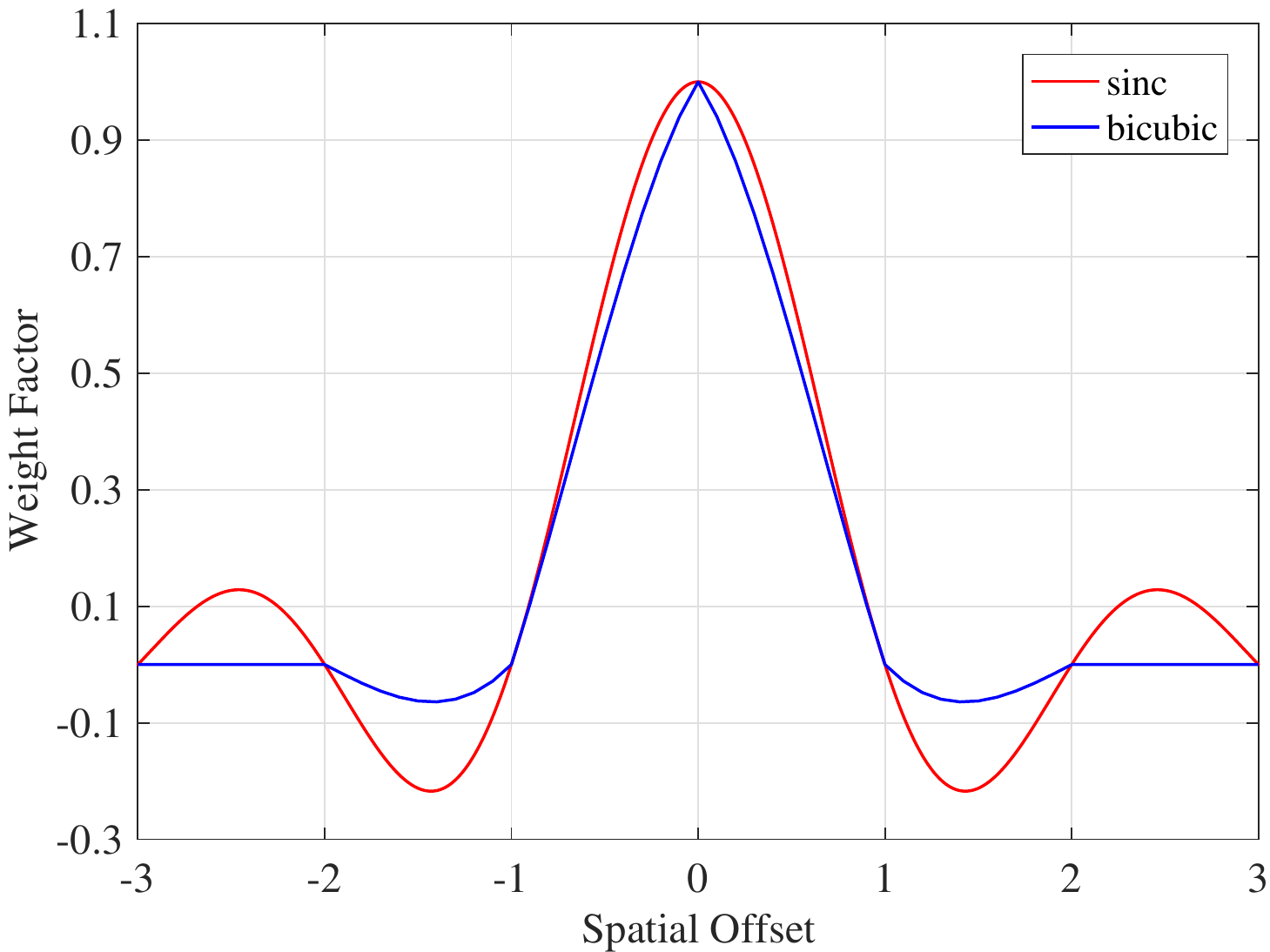}
  \end{minipage}}
  \subfigure[]{\label{frequency_domain}
  \begin{minipage}[t]{0.23\textwidth}
    \centering
    \includegraphics[scale = 0.28]{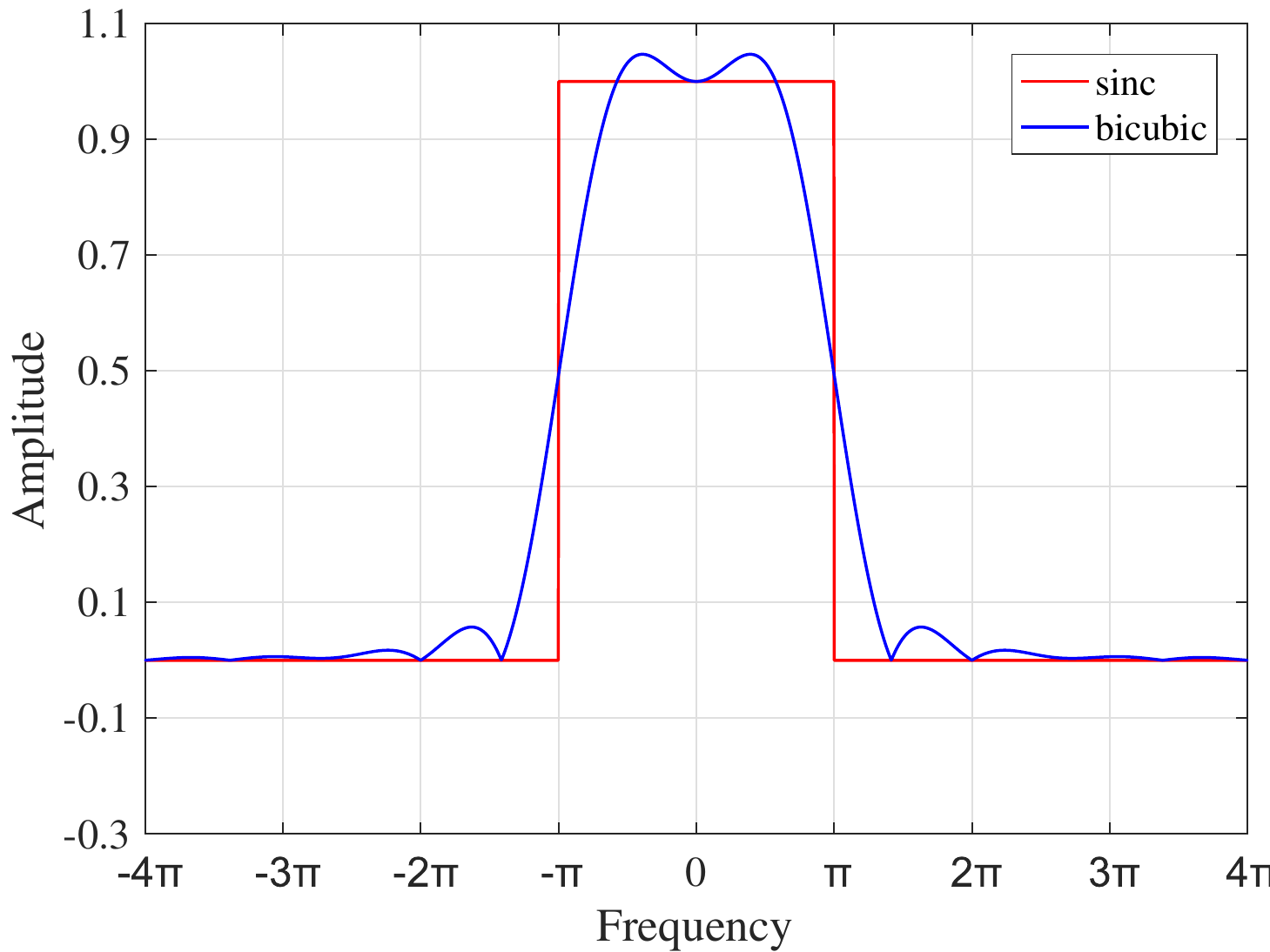}
  \end{minipage}}
  \vspace{-2mm}
  \caption{Graphical comparison between bicubic downsampling and $k$-space truncation in spatial domain (a) and frequency domain (b). The truncation in frequency domain corresponds to a sinc function in spatial domain (1D).}
  \label{fig:image_degradation}
\end{figure}

As in \cite{Zhao2018Channel}, we also investigate two image degradation model in this work, i.e., the bicubic downsampling and the $k$-space truncation. The truncation degradation can be considered as more aggressive because the information outside the sampling range is ``steeply'' cut off without any cushion (Fig.\ref{fig:image_degradation}). As mentioned above, it can be verified by the fact that the bicubic interpolation performs better in bicubic degradation than in truncation degradation. However, the performance of the last few models (e.g., CSN and CSSFN) in Table \ref{tab:4} and Table \ref{tab:5} is contrary to that of bicubic interpolation. On the other hand, the truncation degradation simulates the real MRI acquisition that operates in $k$-space and truncates the frequency spectrum of the object. This indicates that these models are more suitable for the scenarios of MR image super-resolution.

\section{Conclusion}
\label{sec:conclusion}
Channel discrimination is an effective way to improve the SR performance of deep models in the context of degradation of training samples, but how to fuse the channel information remains to be explored. We present a serial fusion strategy for channel discrimination in this work. The hierarchical features are first divided into several ``small'' subfeatures along channel direction, which are then integrated into a single branch in a serial manner. The proposed CSSFN model has the channel discrimination ability in that each subfeature is processed in different network depths. To further improve the information flow of the network, we combine serial fusion with a DGFF to integrate the intermediate features. Extensive quantitative and qualitative experiments demonstrate that our CSSFN model achieves superiority over other state-of-the-art SISR methods. Besides, the serial fusion mode might shed some light on the development of other information fusion methods and channel discrimination methods.


\ifCLASSOPTIONcaptionsoff
  \newpage
\fi

%








\end{document}